\documentclass{article}
\pdfoutput=1

\PassOptionsToPackage{numbers, compress}{natbib}
\usepackage{natbib}

\usepackage[preprint]{neurips_2022}

\usepackage[utf8]{inputenc} 
\usepackage[T1]{fontenc}    
\usepackage{hyperref}       
\usepackage{url}            
\usepackage{booktabs}       
\usepackage{amsfonts}       
\usepackage{nicefrac}       
\usepackage{microtype}      
\usepackage{xcolor}         
\usepackage{wrapfig}

\usepackage{adjustbox}
\usepackage{graphics}
\usepackage{amsmath}
\usepackage{enumitem}
\usepackage[ruled,linesnumbered,vlined]{algorithm2e}
\usepackage{algorithmicx}
\usepackage{algpseudocode}
\usepackage[english]{babel}

\newtheorem{problem}{Problem}
\usepackage{mathabx}
\usepackage{multirow}
\usepackage{subcaption}
\usepackage[normalem]{ulem}
\usepackage{amsmath,bm}
\useunder{\uline}{\ul}{}

\title{Contrastive Graph Few-Shot Learning}

\author{
Chunhui Zhang\textsuperscript{\rm 1}, Hongfu Liu\textsuperscript{\rm 1}, Jundong Li\textsuperscript{\rm 2}, Yanfang Ye\textsuperscript{\rm 3}, Chuxu Zhang\textsuperscript{\rm 1} \\
  \textsuperscript{\rm 1}Brandeis University, USA\\
      \textsuperscript{\rm 2}University of Virginia, USA\\ 
    \textsuperscript{\rm 3}University of Notre Dame, USA\\ 
  \texttt{\{chunhuizhang,hongfuliu,chuxuzhang\}@brandeis.edu} \\
  \texttt{jundong@virginia.edu, yye7@nd.edu} \\
}

\begin{document}

\maketitle

\begin{abstract}
Prevailing deep graph learning models often suffer from label sparsity issue. Although many graph few-shot learning (GFL) methods have been developed to avoid the performance degradation in face of limited annotated data, they excessively rely on labeled data, where the distribution shift in the test phase might result in impaired generalization ability. Additionally, they lack a general purpose as their designs are coupled with task or data-specific characteristics. To this end, we propose a general and effective \textbf{C}ontrastive \textbf{G}raph \textbf{F}ew-shot \textbf{L}earning framework (CGFL). CGFL leverages a self-distilled contrastive learning procedure to boost GFL. Specifically, our model firstly pre-trains a graph encoder with contrastive learning using unlabeled data. Later, the trained encoder is frozen as a teacher model to distill a student model with a contrastive loss. The distilled model is finally fed to GFL. CGFL learns data representation in a self-supervised manner, thus mitigating the distribution shift impact for better generalization and making model task and data-independent for a general graph mining purpose.
Furthermore, we introduce an information-based method to quantitatively measure the capability of CGFL.  Comprehensive experiments demonstrate that CGFL outperforms state-of-the-art baselines on several graph mining tasks in the few-shot scenario. We also provide quantitative measurement of CGFL's success.
 
\end{abstract}

\section{Introduction}
\label{sec:introduction}
Deep graph learning, e.g., graph neural networks (GNNs), has recently attracted tremendous attention due to its remarkable performance in various application domains, such as social/information systems~\cite{kipf2016semi,hamilton2017inductive}, molecular chemistry/biology~\cite{jin2017predicting,hao2020asgn}, and recommendation~\cite{ying2018graph,fan2019graph}. The success of GNNs often relies on massive annotated samples, which contradicts the fact that it is expensive to collect sufficient labels. This motivates the graph few-shot learning (GFL) study  to tackle performance degradation
in the face of limited labeled data.

Previous GFL models are built on meta-learning (or few-shot learning) techniques, either metric-based approaches~\cite{vinyals2016matching,snell2017prototypical} or optimization-based algorithms~\cite{finn2017model}. They aim to quickly learn an effective GNN adapted to new tasks with few labeled samples. GFL has been applied to a variety of graph mining tasks, including node classification~\cite{zhou2019meta,huang2020graph}, relation prediction~\cite{xiong2018one,lv2019adapting,zhang2020few}, and graph classification~\cite{chauhan2020few,ma2020adaptive}. Despite substantial progress, most previous GFL models still have the following limitations:
\textit{(i) Impaired generalization.} Existing GFL methods excessively
rely on labeled data and attempt to inherit a strong inductive bias for new tasks in the test phase. However, a distribution shift exists between non-overlapping meta-training data and meta-testing data. Without supervision signals from ground-truth labels, GFL may not learn an effective GNN for novel classes of test data. This gap limits the meta-trained GNN's generalization and transferability. 
\textit{(ii) Constrained design.} Most of the current GFL methods lack a general purpose as they possess the premise that the designated task is universally the same prior across different graph tasks or datasets, which in fact is not always guaranteed. For example,
GSM~\cite{chauhan2020few} needs to manually define a superclass of graphs, which cannot expand to node-level tasks. The task or data-specific design limits the GFL's utility for different graph mining tasks.

The above challenges call for a new generic GFL framework that can learn a generalizable, transferable, and effective GNN for various graph mining tasks with few labels. {Fortunately, contrastive learning has emerged to alleviate the dependence on labeled data, and learn label-irrelevant but transferable representations from unsupervised pretext tasks for vision, language, and graphs~\cite{chen2020simple, gao-etal-2021-simcse,you2020graphcl, sohn2020fixmatch}.} Thus, the natural idea is to leverage contrastive learning to boost GFL.

In this work, we are motivated to develop a general and effective \textbf{C}ontrastive \textbf{G}raph \textbf{F}ew-shot \textbf{L}earning framework (CGFL) with contrastive learning. 
To be specific, the proposed framework firstly pre-trains a GNN by minimizing the contrastive loss between two views' embeddings generated in two augmented graphs. Later, we introduce a self-distillation step to bring additional elevation: the pre-trained GNN is frozen as a teacher model and kept in the contrastive framework to distill a randomly initialized student model by minimizing the agreement of two views' embeddings generated by two models. Both pre-training and the distillation steps can work at the meta-training and meta-testing phases without requiring labeled data. Finally,
the distilled student model is taken as the initialized model fed to GFL for few-shot graph mining tasks. CGFL pre-trains GNN self-supervised, thus mitigating the negative impact of distribution shift. The learned graph representation is transferable and discriminable for new tasks in the test data. Besides, our simple and generic framework of CGFL is applicable for different graph mining tasks. Furthermore, to quantitatively measure the
capability of CGFL, we introduce information-based method to measure the quality of learned node (or graph) embeddings on each layer of the model: we allocate each node a learnable variable as a noise and train these variables to maximize the entropy while keeping the change of output as small as possible. 

To summarize, our contributions in this work are:
\begin{itemize}[leftmargin=*]
\vspace{-0.15in}
\item We develop a general and effective framework named CGFL to leverage a self-distilled contrastive learning procedure to boost GFL. CGFL mitigates distribution shift impact and has the task and data-independent capacity for a general graph mining purpose. 
\vspace{-0.05in}
\item We introduce an information-based method to quantitatively measure the capability of CGFL by measuring the quality of learned node (or graph) embeddings. To the best of our knowledge, this is the first study to explore GFL model measurement. 
\vspace{-0.05in}
\item Comprehensive experiments on multiple graph datasets demonstrate that CGFL outperforms state-of-the-art methods for both node classification and graph classification tasks in the few-shot scenario. Additional measurement results further show that CGFL learns better node (or graph) embeddings than baseline methods. 
\end{itemize}

\section{Related Work}
\label{sec:related_work}
\textbf{Few-Shot Learning on Graphs}. 
Many GFL models~\citep{zhang2022few} have been proposed to solve various graph mining problems in face of label sparsity issue, such as node classification~\citep{yao2020graph,ding2020graph,huang2020graph,qian2021adapting,qian2021distilling,wang2022task,zhang2022few}, relation prediction~\cite{xiong2018one,lv2019adapting,chen2019meta,zhang2020few,zhang2020few2}, and graph classification~\cite{chauhan2020few,ma2020adaptive,guo2021few,wang2021property}. They are built on meta-learning (or few-shot learning) techniques that can be categorized into two major groups: (1) metric-based approaches~\cite{vinyals2016matching,snell2017prototypical}; (2) optimization-based algorithms~\cite{finn2017model}. For
the first group, they learn effective similarity
metric between few-shot support data and query data. For example, 
GPN~\cite{ding2020graph} conducts node informativeness propagation to build weighted class prototypes for a distance-based node classifier. 
The second group proposes to learn well-initialized GNN parameters that can be fast adapted to new graph tasks with few labeled data. For instance,
G-Meta~\cite{huang2020graph} builds local subgraphs to extract subgraph-specific information and optimizes GNN via MAML~\cite{finn2017model}. Unlike prior efforts that rely on labeled data and have the task and data-specific design, we aim to build a novel framework that explores unlabeled data and has a generic design for a general graph mining purpose.
\\
\textbf{Self-Supervised Learning on Graphs}. 
Recently, self-supervised graph learning (SGL) has attracted significant attention due to its effectiveness in pre-training GNN and competitive performance in various graph mining applications. 
{Previous SGL models can be categorized into two major groups: generative or contrastive, according to their learning tasks~\cite{SSLSurveyGorC, sohn2020fixmatch}.} The generative models learn graph representation by recovering feature or structural information on the graph. The task can solely recover adjacency matrix alone~\cite{GraphRNN} or along with the node features~\cite{GPT-GNN}. 
As for the contrastive methods, they firstly define the node context which can be node-level or graph-level instances. Then, they perform contrastive learning by either maximizing the mutual information between the node-context pairs~\cite{MVGRL,DGI,InfoGraph} or by discriminating context instances~\cite{GCC,GCA,zhao2021multi,yu2022sail}. {In addition to above strategy, recently random propagation applies graph augmentation~\cite{Rong2020DropEdge:} for semi-supervised learning~\cite{NEURIPS2020_fb4c835f}.} Motivated by the success of SGL, we propose to leverage it to boost GFL. 

\section{Preliminary}
\label{sec:preliminary}
\textbf{GNNs}.
A graph is represented as $G = (V, E, X)$,  where $V$ is the set of nodes, $E \subseteq  V \times V$ is the set of edges, and $X$ is the set of node attributes. GNNs~\cite{hamilton2017inductive,xu2018powerful} learn compact representations (embeddings) by considering
both graph structure $E$ and node attributes $X$. To be specific, let  $f_\theta(\cdot)$ denote a GNN encoder with parameter $\theta$, the updated embedding of node $v$ at the $l$-th layer of GNN can be formulated as:
\begin{equation}
    h_v^{(l)} = \mathcal{M}(h_{v}^{(l-1)}, \{h_{u}^{(l-1)}~|~\forall u \in \mathcal{N}_v\};\theta), 
\label{eq:gnn-node-i}
\end{equation}
where $\mathcal{N}_v$ denotes the neighbor set of $v$; $\mathcal{M}(\cdot)$ is the message passing function for neighbor information aggregation, such as a mean pooling layer followed by a fully-connected (FC) layer; $h_v^{(0)}$ is initialized with node attribute $X_{v}$. The whole graph embedding can be computed over all nodes' embeddings as:
\begin{equation}
    h_G^{(l)} = \textrm{READOUT}\{h_{v}^{(l)}~|~\forall v \in V\}, 
\label{eq:gnn-graph}
\end{equation}
where the READOUT function can be a simple permutation invariant function such as summation. 

\textbf{GFL Setting and Problem}. Let $\mathcal{C}_{base}$ and $\mathcal{C}_{novel}$ denote the base classes set and novel (new) classes set in training data $\mathcal{T}_{train}$ and testing data $\mathcal{T}_{test}$, respectively. 
Similar to the general meta-learning problem~\cite{finn2017model}, the purpose of graph few-shot learning (GFL) is to train a GNN encoder $f_\theta(\cdot)$ over $\mathcal{C}_{base}$, such that the trained GNN encoder can be quickly adapted to $\mathcal{C}_{novel}$ with few labels per class. Note that there is no overlapping between base classes and novel classes, i.e., $\mathcal{C}_{base} \cap \mathcal{C}_{novel} = \emptyset$. 
In $K$-shot setting, during the meta-training phase,
a batch of classes (tasks) is randomly sampled from $\mathcal{C}_{base}$, where $K$ labeled instances per class are sampled to form the support set $\mathcal{S}$ for model training and the remaining instances are taken as the query set $\mathcal{Q}$ for model evaluation. After sufficient training, the model is further transferred to the meta-testing phase to conduct $N$-way classification over $\mathcal{C}_{novel}$ ($N$ is the number of novel classes), where each class is only with $K$ labeled instances. GFL applies to different graph mining problems, depending on the class meaning. Each class corresponds to a node label for the node classification problem or corresponds to a graph label for the graph classification problem. In this work, we will study both node classification and graph classification problems under few-shot setting, which are formally defined as follows:
\begin{problem}
\textbf{Few-Shot Node Classification}. Given a graph $G = (V, E, X)$ and labeled nodes of $\mathcal{C}_{base}$, the problem is to learn a GNN $f_\theta(\cdot)$ to classify nodes of $\mathcal{C}_{novel}$, where each class in $\mathcal{C}_{novel}$ only has few labeled nodes. 
\end{problem}

\begin{problem}
\textbf{Few-Shot Graph Classification}. Given a set of graphs $\mathcal{G}$ and labeled graphs of $\mathcal{C}_{base}$, the problem is to learn a GNN $f_\theta(\cdot)$ to classify graphs of  $\mathcal{C}_{novel}$, where each class in $\mathcal{C}_{novel}$ only has few labeled graphs.  
\end{problem}

Unlike previous studies that rely on labeled data of $\mathcal{T}_{train}$ and $\mathcal{T}_{test}$ for GFL model training and adaption, we consider both unlabeled graph information and labeled data to learn GFL model for solving the above problems.

\section{Methodology}
\label{sec:model}
\begin{figure*}[!htb]
\centering
\includegraphics[width=1.0\linewidth]{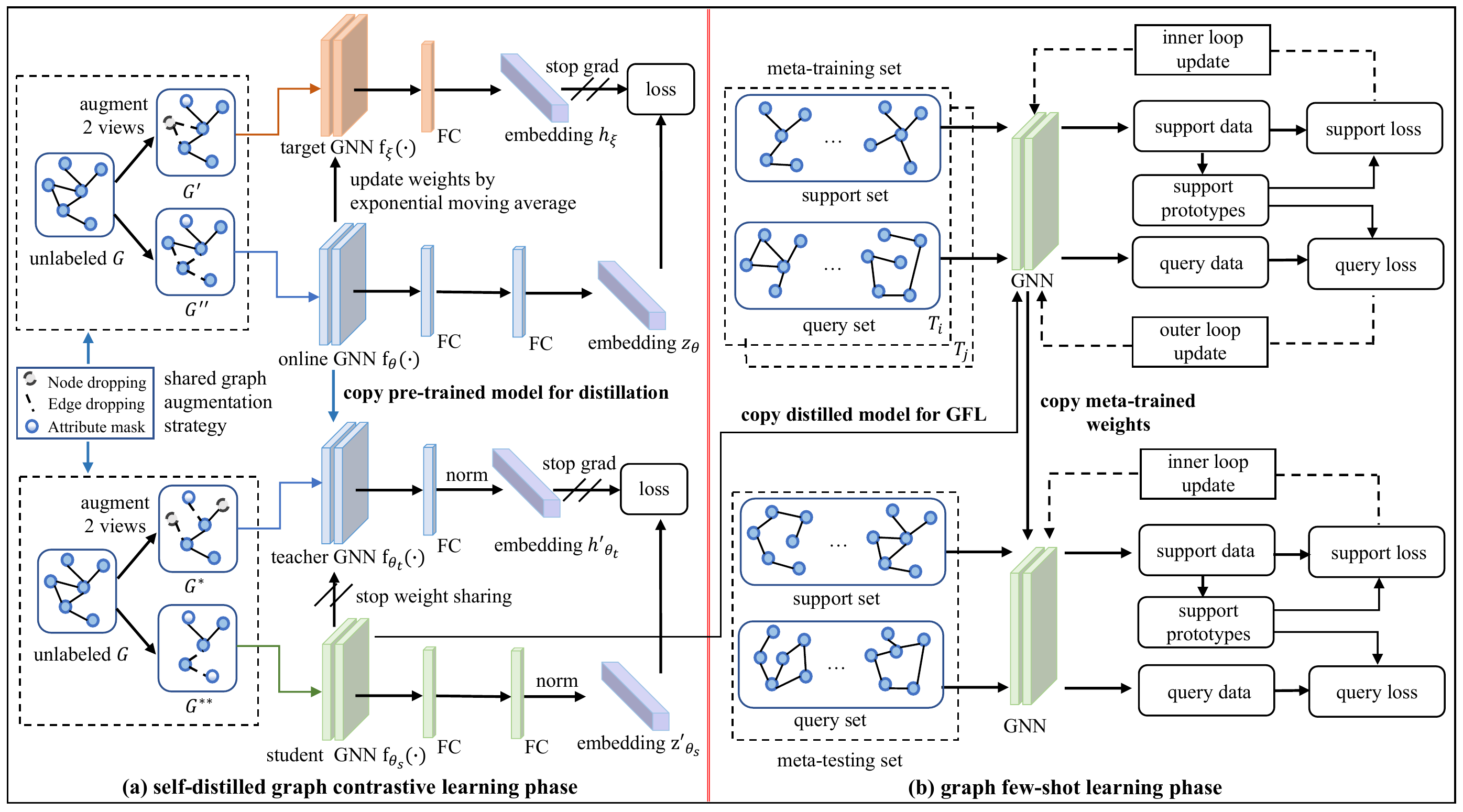}
\vspace{-0.2in}
\caption{The overall framework of CGFL: (a) self-distilled graph contrastive learning phase which pre-trains a GNN encoder with contrastive learning and further evaluates the model with knowledge distillation in self-supervised manner; (b) graph few-shot learning phase which takes the distilled student network as the initialized model and employs meta-learning algorithm for model optimization.}
\label{fig:ssl}
\vspace{-0.2in}
\end{figure*}

Figure~\ref{fig:ssl} illustrates the proposed CGFL framework, which includes two phases:  self-distilled graph contrastive learning and graph few-shot learning (GFL). In the first phase (Figure~\ref{fig:ssl}(a)), the framework pre-trains a GNN encoder with contrastive learning, then introduces knowledge distillation to elevate the pre-trained GNN in a self-supervised manner. The distilled GNN is finally fed to the GFL phase (Figure~\ref{fig:ssl}(b)) for few-shot graph mining tasks. In addition to the proposed framework, we introduce an information-based method to measure the superiority of CGFL quantitatively.

\subsection{Self-Distilled Graph Contrastive Learning}
\textbf{GNN Contrastive Pre-training}. In the first phase, we firstly introduce contrastive learning to pre-train GNN. Inspired by the representation bootstrapping technique~\cite{grill2020bootstrap}, our method learns node (or graph) representation by discriminating context instances. Specifically, two GNN encoders: an online GNN $f_\theta(\cdot)$ and a target GNN $f_\xi(\cdot)$, are introduced to encode two randomly augmented views of a given graph. The online GNN is supervised under the target GNN's output, while the target GNN is updated by the online GNN's exponential moving average. The contrastive pre-training step is shown in Figure~\ref{fig:ssl}(a).
\\
\textit{Graph Augmentation:} The given graph $G$ is processed with randomly data augmentations to generate a contrastive pair ($G^{\prime}$, $G^{\prime\prime}$) as the input for two GNN branches (online branch and target branch) of the following GNN training. In this work, we apply a combination of stochastic node feature masking, edge removing, and node dropping with constant probabilities for graph augmentation.
\\
\textit{GNN Update:} With the generated graph pair ($G^{\prime}$, $G^{\prime\prime}$), the online GNN $f_\theta(\cdot)$ and the target GNN $f_\xi(\cdot)$ are respectively utilized to process $G^{\prime}$ and $G^{\prime\prime}$ for node (or graph) embeddings generation. Both GNNs have the same architecture, while a two-layer FC (one-layer FC) is attached after online GNN (target GNN) to refine embedding. The reason that two branches have different FC layers is to prevent the prediction of the online model from being exactly the same as the output of the target model, thus avoiding the learned representation collapse. Later, to enforce online GNN's embeddings $z_\theta$ approximate the target GNN's embeddings $h_\xi$, the mean squared error between them is formulated as the objective function:
\begin{equation}
    \mathcal{L}_{\theta,\xi} = {\lVert {z_{\theta} - h_{\xi}}\rVert}_{2}^{2}=2-2\cdot \frac{{z_{\theta}, h_{\xi}}}{{\lVert {z_{\theta} }\rVert}_{2} \cdot {\lVert {h_{\xi}}\rVert}_{2}}.
\label{eq:cosine-loss}
\end{equation}
The parameters $\theta$ of online GNN are updated with Adam optimizer~\cite{adam15}:
\begin{equation}
    \theta \leftarrow \textrm{Adam}(\theta, \nabla_\theta \mathcal{L}_{\theta,\xi}, \eta),
\label{eq:update-online}
\end{equation} 
where $\eta$ is the learning rate.  The target GNN provides the regression target to supervise the online GNN, and its parameters $\xi$ are updated as exponential moving average (EMA) of the online GNN parameters $\theta$. More precisely, $\xi$ is updated as follows:
\begin{equation}
    \xi \leftarrow \tau\xi + (1-\tau)\theta,
\label{eq:update-target}
\end{equation}
where $\tau \in [0, 1]$ is the decay rate. Note that the target GNN stops the backpropagation from $\mathcal{L}_{\theta,\xi}$, and it is only updated by EMA.
\textbf{Contrastive Distillation}.
With the pre-trained GNN $f_\theta(\cdot)$ obtained in the previous step, we introduce a self-distillation step to elevate $f_\theta(\cdot)$. This is inspired by the Born-again strategy~\cite{furlanello2018born}, which implies a well-trained teacher can boost a random initialized identical student. The distillation step adopts a similar contrastive framework as the previous step, as shown in Figure~\ref{fig:ssl}(a). Specifically, we load the pre-trained GNN $f_\theta(\cdot)$ and take it as the teacher model $f_{\theta_{t}}(\cdot)$. The teacher model is frozen and applied to distill a student model $f_{\theta_{s}}(\cdot)$.
Two augmented views ($G^{*}$, $G^{**}$) of a graph $G$ are generated and fed to $f_{\theta_{t}}(\cdot)$ and $f_{\theta_{s}}(\cdot)$, respectively. Later,
the student's normalized output is forced to approximate the teacher's normalized output as follows:
\begin{equation}
    \mathcal{L}_{\theta_s} = \lVert z^\prime_{\theta_s} - h^\prime_{\theta_t}\rVert_2^2=2-2\cdot \frac{{z^\prime_{\theta_s}, h^\prime_{\theta_t}}
}{\lVert z^\prime_{\theta_s} \rVert_2 \cdot \lVert h^\prime_{\theta_t}\rVert_2},
\label{eq:cosine-loss-kd}
\end{equation}
\begin{equation}
    z^\prime_{\theta_s} = \frac{z_{\theta_s}}{\lVert z_{\theta_s}\rVert_2},~h^\prime_{\theta_t} = \frac{h_{\theta_t}}{\lVert h_{\theta_t}\rVert_2},
\label{eq:cosine-loss-kd-norm}
\end{equation}
where $z_{\theta_s}$ and $h_{\theta_t}$ are teacher's output embeddings and student's output embeddings, respectively. The student model is updated as follows:
\begin{equation}
    \theta_{s} \leftarrow \textrm{Adam}(\theta_{s}, \nabla_{\theta_{s}} \mathcal{L}_{\theta_{s}}, \eta).
\label{eq:stu-update-online}
\end{equation} 
Different from EMA (Eqn.~\ref{eq:update-target}) for target GNN update in contrastive pre-training, the teacher model is frozen and can be seen as a special case of EMA:
\begin{equation}
    \theta_t \leftarrow \tau\theta_t + (1-\tau)\theta_s,~\tau=1.
\label{eq:update-target-kd}
\end{equation}

\subsection{Graph Few-Shot Learning}
In GFL phase, we take the distilled student GNN $f_{\theta_{s}}(\cdot)$ generated in the former phase as the initialized GNN model and employ the optimization-based algorithm, i.e., model-agnostic meta-learning (MAML)~\cite{finn2017model}, to train the model for few-shot graph mining tasks. 
During meta-training, for task $T_i$, the task specific parameters $\theta_{{s, i}}'$ is computed using a number of gradient descent updates over the support set $\mathcal{S}_{i}$ of $T_i$  (i.e., inner-loop):
\begin{equation}
    \theta_{{s, i}}' \leftarrow \theta_s - \alpha\nabla_{\theta_s}\mathcal{L}^{\mathcal{S}_{i}}_{\mathcal{T}_i}(\theta_s), 
\label{eq:maml-1}
\end{equation}
where $\alpha$ is the learning step size, $\mathcal{L}^{\mathcal{S}_{i}}_{\mathcal{T}_i}$ denotes the downstream task loss over $\mathcal{S}_{i}$. In this work, we employ prototypical loss~\cite{snell2017prototypical} for node or graph classification, which uses embeddings extracted from the support set by a neural network as the class prototype, and the query set is classified according to the distance between its embeddings and prototypes. The task-specific parameter $\theta_{{s, i}}'$ is further utilized to compute the loss over query set $\mathcal{Q}_{i}$ of $T_i$: $\mathcal{L}^{\mathcal{Q}_{i}}_{\mathcal{T}_i}(f_{\theta_{{s, i}}'})$. Later, $\mathcal{L}^{\mathcal{Q}_{i}}_{\mathcal{T}_i}(f_{\theta_{{s, i}}'})$ of a batch of randomly sampled tasks are summed up to update the model parameters $\theta_{s}$ (i.e., outer-loop):
\begin{equation}
\theta_s \leftarrow \theta_s - \beta \nabla_{\theta_s} \sum_i \mathcal{L}^{\mathcal{Q}_{i}}_{\mathcal{T}_i}(f_{\theta_{{s, i}}'}),
\label{eq:maml-2}
\end{equation}
where $\beta$ is the learning step size. During meta-testing, the same procedure above is applied using the final meta-updated parameter $\theta^*_{s}$ for novel tasks (without outer-loop). 
In particular, $\theta^*_{s}$ is learned from
knowledge across meta-training tasks and is the optimal parameter to quickly adapt to novel tasks. Note that the GFL algorithm can be applied to different graph mining problems by changing the task meaning, i.e., each task corresponds to a node class (or graph class) for node classification (or graph classification).

\subsection{Quantitative Measurement of GFL}
\label{sec:inforamtion-loss}
The previous GFL studies target developing better methods in performance while none of them has thought about model capability measurement. 
In light of this, we extend the measurement of neural network model~\cite{ma2019quantifying} to graph data, and  quantitatively show why different GFL models can learn node (or graph) representations at different extents. 
The proposed method provides a measurement of information encoded in GNN for the input graph. Specifically, let $Z$ denote GNN hidden state of a GFL model, the information of the input graph $G$ encoded by $Z$ can be measured by mutual information $MI(G;Z)$: 
\begin{equation}
MI(G;Z)=H(G)-H(G|Z),
\label{eq:infoloss-1}
\end{equation}
where $H(\cdot)$ denotes the entropy; $H(G)$ is a constant; $H(G|Z)$ represents the amount of discarded information after $G$ is processed by GNN and encoded by $Z$. We can compute $H(G|Z)$ by decomposing it into the node level: 
\begin{equation}
H(G|Z) = \int_{{\bf z}\in Z}p({\bf z})H(G|{\bf z})d{\bf z},
\label{eq:infoloss-2}
\end{equation}
where $\textbf{z} = f(x)$ denotes GNN hidden state corresponding to attribute $x$ of a node. 
$H(G|{\bf {z}})$ reflects how much information from $x$ is discarded by {\bf z} during the forward propagation. 
To disentangle information components
of individual nodes from the whole graph, we assume that each node is independent of each other and have:
\begin{equation}
H(G|{\bf {z}}) = \sum_{i}H(x_{i}|{\bf {z}}),
\end{equation}
where $x_{i}$ denotes a random variable of $i$-th node attribute in the graph. 
Then, we introduce a  noise perturbation-based method to approximate $H(x_{i}|{\bf {z}})$. Specifically,  let $\widetilde{x_i}=x_i+{\bf \epsilon}_i$ (${\bf \epsilon}_i \sim \mathcal{N}(0, {\bf \Sigma}_i={\bf \sigma}^2_i{\bf I})$) and we aim to optimize the following loss function:
\begin{equation}
\mathcal{L}(\bm{\sigma})=\mathbb{E}_{\bf \epsilon}\lVert f(\widetilde{x}_i)-{\bf z} \lVert^2-\beta\sum_{i=1}^{n}H(\widetilde{ x}_i|{\bf z})|_{\epsilon_i \sim \mathcal{N}(0, \sigma_i^2{\bf I})},
\label{eq:infoloss-6}
\end{equation}
where $\bm{\sigma}$ = $[\sigma_1, \sigma_2, \cdots, \sigma_n]$ are learnable parameters; 
$\beta$ is trade-off weight. In particular, the first term of the above objective minimizes difference between the encoded embedding of noisy input and the hidden state while the second term encourages a high conditional entropy according to the maximum entropy principle. In this way, we have $p(\widetilde{x_i}|{\bf z})=p({\bf \epsilon}_i)$ and $H(x_i|{\bf z})$ is approximated by $H(\widetilde{x}_i|{\bf z})$:
\begin{equation}
H(\widetilde{x_i}|{\bf z})= p( \widetilde{x_i}|{\bf z})\log_{}p(\widetilde{x_i}|{\bf z}) \propto  \log \sigma_i + C, 
\label{eq:infoloss-7}
\end{equation}
where $C = \frac{1}{2}\log (2\pi e)$. By taking Eqn.~\ref{eq:infoloss-7} into Eqn.~\ref{eq:infoloss-6}, 
we can optimize $\mathcal{L}(\bm{\sigma})$ with Adam optimizer and obtain the optimal $\bm{\sigma}$ for computing the overall discarded information $H(G|Z)$. Furthermore, we can explain GFL model's capability by comparing discarded information of different models. Ideally, we expect the GFL model to encode valid node (or graph) embeddings as much as possible and therefore discard information as small as possible.

\section{Experiments}
\label{sec:experiment}
We conduct extensive experiments on multiple graph datasets to evaluate the model performance compared with state-of-the-art models. We first describe experimental settings and then discuss the performance comparison of different models. Finally, discarded information is computed to show the capabilities of different GFL models. More experimental results are provided in Appendix~\ref{app: results}. 

\subsection{Experimental Setup}
\label{sec:exp-set}
\textbf{Datasets}.
We use multiple graph datasets to conduct experiments: for the node classification task, we use ogbn-arxiv~\citep{hu2020open}, Tissue-PPI~\citep{hamilton2017inductive}, Fold-PPI~\citep{zitnik2017predicting}, Cora~\citep{sen2008collective}, and Citeseer~\citep{sen2008collective}; for the graph classification task, we use datasets in ~\citep{chauhan2020few}, i.e., Letter-High, Triangles, Reddit-12K, and Enzymes. The detailed information of datasets is illustrated in Appendix~\ref{app: dataset}. 
\\
\textbf{Baseline Methods}. 
We employ a variety of baseline methods for model comparison in two tasks. For few-shot node classification, we use node2vec~\citep{grover2016node2vec}, DeepWalk~\citep{perozzi2014deepwalk}, Meta-GNN~\citep{zhou2019meta}, FS-GIN~\citep{xu2018powerful}, FS-SGC~\citep{wu2019simplifying}, No-Finetune~\citep{triantafillou2019meta},
Finetune~\citep{triantafillou2019meta},
KNN~~\citep{triantafillou2019meta},  ProtoNet~\citep{snell2017prototypical}, MAML~\citep{finn2017model}, G-Meta~\citep{huang2020graph}, and TENT~\citep{wang2022task}. For few-shot graph classification, we utilize WL~\citep{shervashidze2011weisfeiler}, Graphlet~\citep{shervashidze2009efficient}, AWE~\citep{ivanov2018anonymous}, Graph2Vec~\citep{narayanan2017graph2vec}, Diffpool~\citep{lee2019self}, CapsGNN~\citep{xinyi2018capsule}, GIN~\citep{xu2018powerful}, GIN-KNN~\citep{xu2018powerful}, GSM-GCN~\citep{chauhan2020few}, GSM-GAT~\citep{chauhan2020few}, and AS-MAML~\citep{ma2020adaptive}. Details of baselines are illustrated in Appendix~\ref{app: implementation}. 
\\
\textbf{Experimental Settings}.
In CGFL, we use GCN~\citep{kipf2016semi} as the GNN backbone for the node classification task, including two-layer graph convolution and one-layer FC. The graph classification task adds an extra average pooling operation as a readout layer. 
In the pretraining phase, we pre-train GNN on the unlabeled graph dataset with contrastive learning. 
For graph data augmentation, the node drop rate is 15\%, the edge removing rate is 15\%; and the feature masking rate is 20\%. The mini-batch size is set to 2,048, and the learning rate is set to 0.05 with a decay factor = 0.9. Meanwhile, the $\tau$ in exponential moving average is 0.999. We consider both inductive and transductive settings. For the inductive setting (\textbf{CGFL-I}), we only use unlabeled data in the training set; for the transductive setting (\textbf{CGFL-T}), we use unlabeled data in both training data and testing data. Additionally, for pre-trained models without knowledge distillation elevation, we refer to them as \textbf{Teacher-I} and \textbf{Teacher-T} for two settings, respectively. 
In GFL phase, we employ MAML to fine-tune the model.
We implement CGFL by PyTorch and train it on NVIDIA V100 GPUs. 
The code is in the supplementary material. 

\subsection{Few-Shot Node Classification}
\textbf{Overall Performance}.
The performances of all models for 3/5-shot node classification 
are reported in Table~\ref{tab:node-cls-1}. According to this Table, we have several findings: 
(1) CGFL outperforms all baseline methods on all datasets, showing the superiority of our model for few-shot node classification;
(2) The improvement of CGFL-I over baseline methods ranges from 1.1\% to 50\% (3-shot) and from 2.1\% to 51\% (5-shot). Simultaneously, this value of CGFL-T ranges from 4.8\% to 55.0\% (3-shot) and from 4.5\% to 56.6\% (5-shot). The significant improvement demonstrates the effectiveness of contrastive pre-training and self-distillation in learning rich node representations from unlabeled graph data and  alleviating the label-hungry issue;
(3) CGFL-T (or Teacher-T) is better than CGFL-I (or Teacher-I). It indicates that the model gets benefits from unlabeled data in the meta-testing set, thus improving generalization ability over testing  data;
{(4) CGFL gains an additional boost compared with the pre-trained teacher model without any label cost in distillation, showing the effectiveness of contrastive distillation.}
\begin{table*}[t]
\vspace{-0.2in}
\caption{Few-shot node classification results. The best results are highlighted in bold while the best baseline results are underlined. -I and -T denote inductive and transductive settings of CGFL, respectively. Teacher indicates the CGFL model without knowledge distillation.}
\vspace{-0.1in}
\renewcommand\arraystretch{1}
\centering
\resizebox{1\textwidth}{!}{
\begin{tabular} {c|cc|cc|cc|cc|cc}
\toprule
\multirow{2}{*}{Method} &\multicolumn{2}{c|}{Tissue-PPI} &\multicolumn{2}{c|}{Fold-PPI} &\multicolumn{2}{c|}{Cora} &\multicolumn{2}{c|}{Citeseer} & \multicolumn{2}{c}{ogbn-arxiv}  \\ 
\cmidrule{2-3} \cmidrule{4-5} \cmidrule{6-7}\cmidrule{8-9} \cmidrule{10-11} & 3-shot & 5-shot & 3-shot & 5-shot & 3-shot & 5-shot & 3-shot & 5-shot  & 3-shot & 5-shot  \\
\midrule
node2vec &48.5$\pm$3.3 &49.3$\pm$3.9 &36.6$\pm$3.7 &37.4$\pm$1.9 &25.7$\pm$1.3 &26.9$\pm$3.0 &20.0$\pm$2.5 &21.7$\pm$2.9&28.9$\pm$4.0 &29.5$\pm$3.7   \\ 
DeepWalk &46.2$\pm$4.8 &47.4$\pm$3.6 &35.0$\pm$4.4 &36.3$\pm$3.2 &25.6$\pm$0.8 &26.7$\pm$2.0 &21.2$\pm$0.6 &22.6$\pm$ 2.7 &30.3$\pm$2.1 &31.5$\pm$3.4 \\ 
Meta-GNN &50.8$\pm$8.1 &53.5$\pm$1.5 &30.8$\pm$5.4 &33.5$\pm$2.1 &\underline{76.8$\pm$0.9} &\underline{79.2$\pm$1.9} &69.4$\pm$1.4 &\underline{72.6$\pm$1.9}&27.3$\pm$1.2 &30.2$\pm$3.6 \\ 
FS-GIN  &49.2$\pm$2.4 &51.5$\pm$3.0 &36.7$\pm$2.1 &39.1$\pm$1.4 &53.5$\pm$1.6 &56.2$\pm$2.8 &50.2$\pm$2.6 &53.2$\pm$3.8 &33.6$\pm$4.2 &36.8$\pm$2.5\\ 
FS-SGC &49.8$\pm$3.8 &52.3$\pm$2.2 &38.0$\pm$1.6 &40.9$\pm$3.9 &57.2$\pm$2.1 &60.3$\pm$1.2 &52.0$\pm$2.1 &54.4$\pm$2.5&34.7$\pm$0.5 &37.3$\pm$1.0  \\ 
No-Finetune &51.6$\pm$0.6 &55.0$\pm$2.1 &37.6$\pm$1.7 &39.9$\pm$3.6 &61.2$\pm$1.2 &64.5$\pm$1.3 &54.9$\pm$1.7 &58.3$\pm$2.5 &36.4$\pm$1.4 &38.8$\pm$2.0 \\ 
Finetune &52.1$\pm$1.3 &54.3$\pm$2.4 &37.0$\pm$2.2 &40.0$\pm$2.6 &63.5$\pm$0.8 &65.7$\pm$2.1 &57.8$\pm$1.8 &59.0$\pm$2.9 &35.9$\pm$1.0 &38.6$\pm$2.5  \\ 
KNN &61.9$\pm$2.5 &65.2$\pm$3.2 &43.3$\pm$3.4 &46.2$\pm$1.9 &67.8$\pm$1.4 &70.3$\pm$3.6 &60.6$\pm$1.4 &63.2$\pm$1.6 &39.2$\pm$1.5 &42.3$\pm$1.8  \\ 
ProtoNet &54.6$\pm$2.5 &57.5$\pm$2.9 &38.2$\pm$3.1 &41.3$\pm$1.1 &42.6$\pm$3.7 &56.6$\pm$2.9 &55.5$\pm$1.5 &58.0$\pm$3.7&37.2$\pm$1.7 &39.7$\pm$1.7   \\ 
MAML &74.5$\pm$5.1 &77.4$\pm$2.7 &48.2$\pm$6.2 &51.3$\pm$3.3 &65.7$\pm$0.9  &68.8$\pm$1.1 &63.1$\pm$1.6  &65.7$\pm$1.7&38.9$\pm$2.1 &41.3$\pm$2.4   \\ 
GPN &\underline{77.3$\pm$3.0} &79.0$\pm$3.6 &57.0$\pm$4.7 &58.2$\pm$3.7 &73.1$\pm$2.4 &76.1$\pm$2.2 &68.3$\pm$1.4 &71.1$\pm$2.0 &44.4$\pm$3.5 &48.2$\pm$4.0   \\
RALE &76.6$\pm$3.3 &79.2$\pm$3.3 &\underline{57.8$\pm$4.5} &58.8$\pm$3.3 &62.8$\pm$3.1 &65.9$\pm$3.2 &\underline{69.9$\pm$2.3} &71.3$\pm$2.2 &45.1$\pm$2.7 &47.8$\pm$1.5 \\
G-Meta &{76.8$\pm$2.9} &\underline{79.4$\pm$2.6} &{56.1$\pm$5.9} &\underline{59.0$\pm$2.5} &{71.9$\pm$2.9} &{74.5$\pm$2.0} &{67.8$\pm$2.2} &{70.8$\pm$3.8} &\underline{45.1$\pm$3.2} &{48.2$\pm$3.1} \\ 
TENT  &- &- &- &- &{64.8$\pm$4.1} &{69.2$\pm$4.5} &{54.2$\pm$3.4} &{62.0$\pm$2.3} &\textbf{55.6$\pm$3.1} &\textbf{62.9$\pm$3.7} \\ 
\midrule
Teacher-I &{77.9$\pm$2.6} &{80.8$\pm$1.7} &{58.8$\pm$3.4} &{61.3$\pm$3.6} &{78.2$\pm$1.3} &{80.9$\pm$1.9}  &{70.0$\pm$1.3} &{72.7$\pm$1.8}&{52.0$\pm$2.0} &{54.9$\pm$1.6}   \\
CGFL-I &{78.7$\pm$2.8} &{81.5$\pm$3.6} &{59.5$\pm$4.1} &{62.0$\pm$2.0}  &{78.5$\pm$1.5} &{81.2$\pm$1.5}  &{70.6$\pm$1.2} &{73.1$\pm$2.0} &{52.8$\pm$1.8} &{55.6$\pm$1.1} \\
Teacher-T &{79.8$\pm$3.1} &{82.9$\pm$1.3} &{63.0$\pm$3.6} &{65.6$\pm$2.2}  &{80.0$\pm$2.7} &{82.6$\pm$1.9}  &{72.1$\pm$1.1} &{75.0$\pm$1.5} &{54.3$\pm$2.7}  &{58.4$\pm$3.9}  \\
CGFL-T &\textbf{80.9$\pm$3.0} &\textbf{84.1$\pm$3.2}  &\textbf{66.9$\pm$3.4} &\textbf{69.0$\pm$2.7} &\textbf{80.7$\pm$1.9} &\textbf{83.5$\pm$3.0} &\textbf{72.6$\pm$1.6} &\textbf{75.3$\pm$2.0}&\textbf{55.2$\pm$2.5} &\underline{58.7$\pm$2.7}  \\ 
\bottomrule
\end{tabular}}
\label{tab:node-cls-1}
\vspace{-0.1in}
\end{table*}
\\
\begin{figure}
\centering
\begin{subfigure}{.5\textwidth}
  \centering
  \includegraphics[width=1\linewidth]{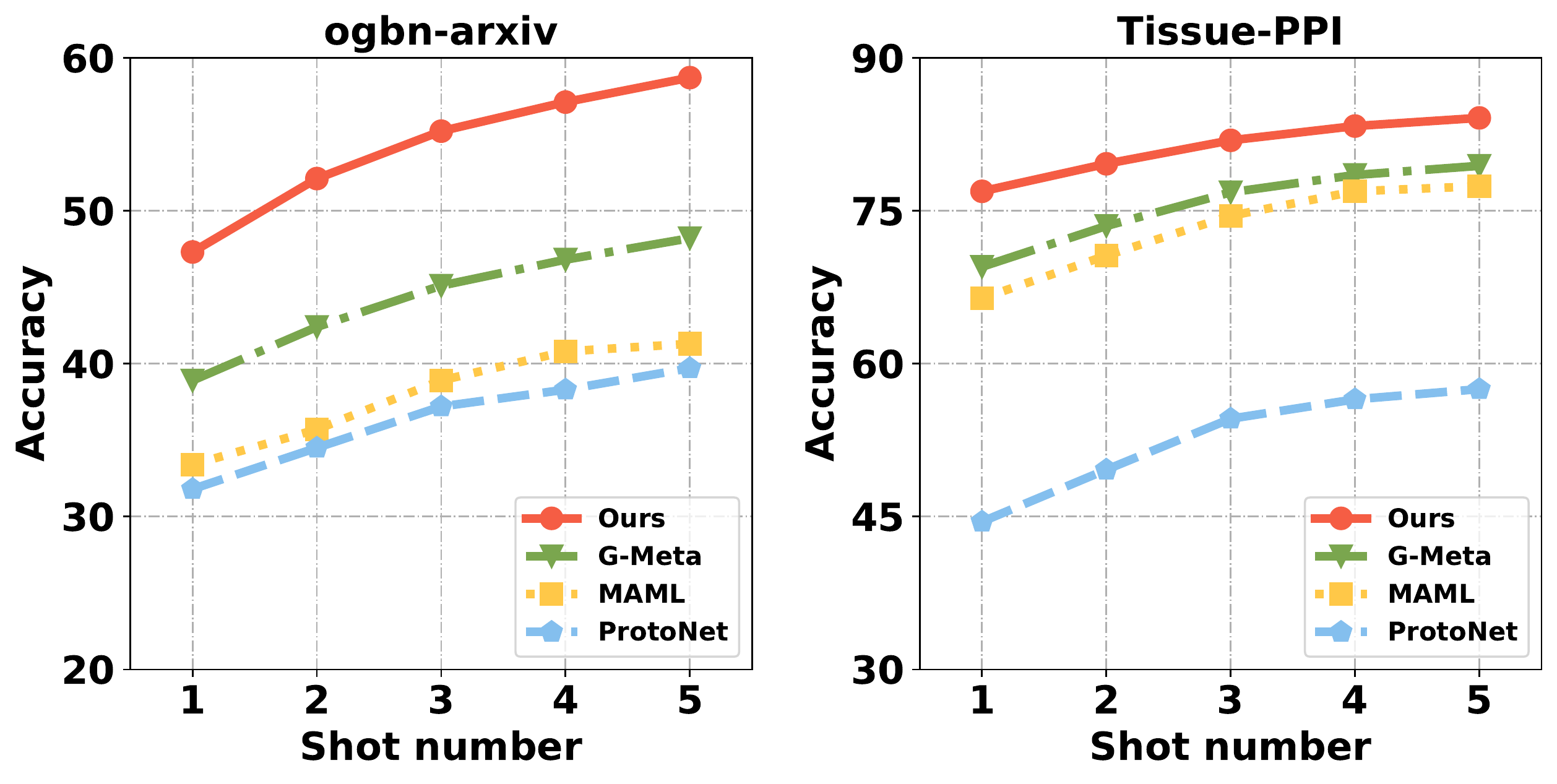}
  \vspace{-0.2in}
  \caption{Impact of shot number.}
\label{fig:node-shot}
\end{subfigure}%
\begin{subfigure}{.5\textwidth}
  \centering
  \includegraphics[width=1\linewidth]{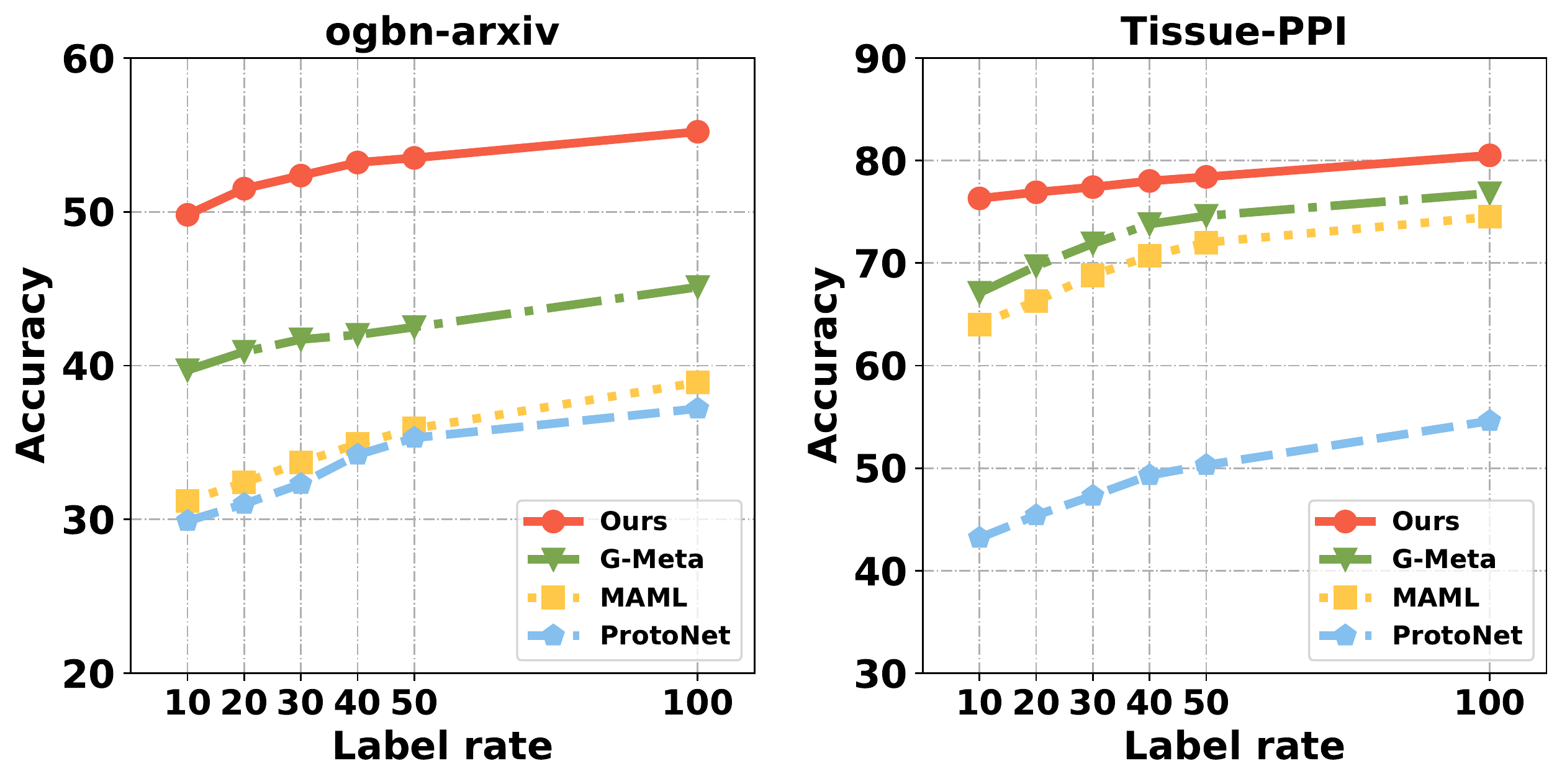}
    \vspace{-0.2in}
  \caption{Impact of label rate.}
  \label{fig:node-lr}
\end{subfigure}
\caption{Impact of shot number and label rate on node classification.}
\vspace{-0.25in}
\label{fig:node-shot-lr}
\end{figure}
\\
\textbf{Impact of Shot Number}.
In Figure~\ref{fig:node-shot}, we show the performance of our model (CGFL-T) under different shot numbers (1 to 5) compared with some selected baselines. It is easy to see that CGFL achieves better accuracy across different shot numbers.
The result demonstrates our model's performance is robust for node classification. Note that we only show results on two datasets and the results on the other datasets (i.e., ogbn-arxiv and Tissue-PPI) are shown in Appendix~\ref{sec:app:fsn}. The same goes for the following analyses. 
\\
\textbf{Impact of Training Label Rate}.
We further evaluate CGFL's performance under different training label rates (10\%, 20\%, 30\%, 40\%, 50\%, 100\%) compared with baseline methods for 3-shot node classification, as shown in Figure~\ref{fig:node-lr}. It is easy to see that (1) CGFL has better results across different label rates; (2) CGFL achieves larger improvement over baseline models when label rate becomes lower (e.g., 10\%), showing contrastive pre-training and self-distillation of CGFL lead to more significant improvement for label sparsity case. 
\\
\textbf{Impact of Data Augmentation}.
As a key step for contrastive learning in CGFL, graph augmentation plays an important role in affecting model performance. Here we conduct experiments to evaluate the model performance with different augmentation strategies. We consider three graph augmentations - node dropping (\textbf{ND}), feature masking (\textbf{FM}), edge removing (\textbf{ER}), and their combinations for CGFL phase and report their performances in Figure~\ref{fig:node-da}. It is easy to find that the combination of three augmentation strategies works better than a single augmentation or the combination of two augmentations. It demonstrates that various graph augmentations are able to generate sufficient contrastive pairs for a better model. 
\\
\begin{minipage}{0.52\textwidth}
\centering
\label{fig:node-aug}
\vspace{0.1in}
    \includegraphics[width=0.9\linewidth]{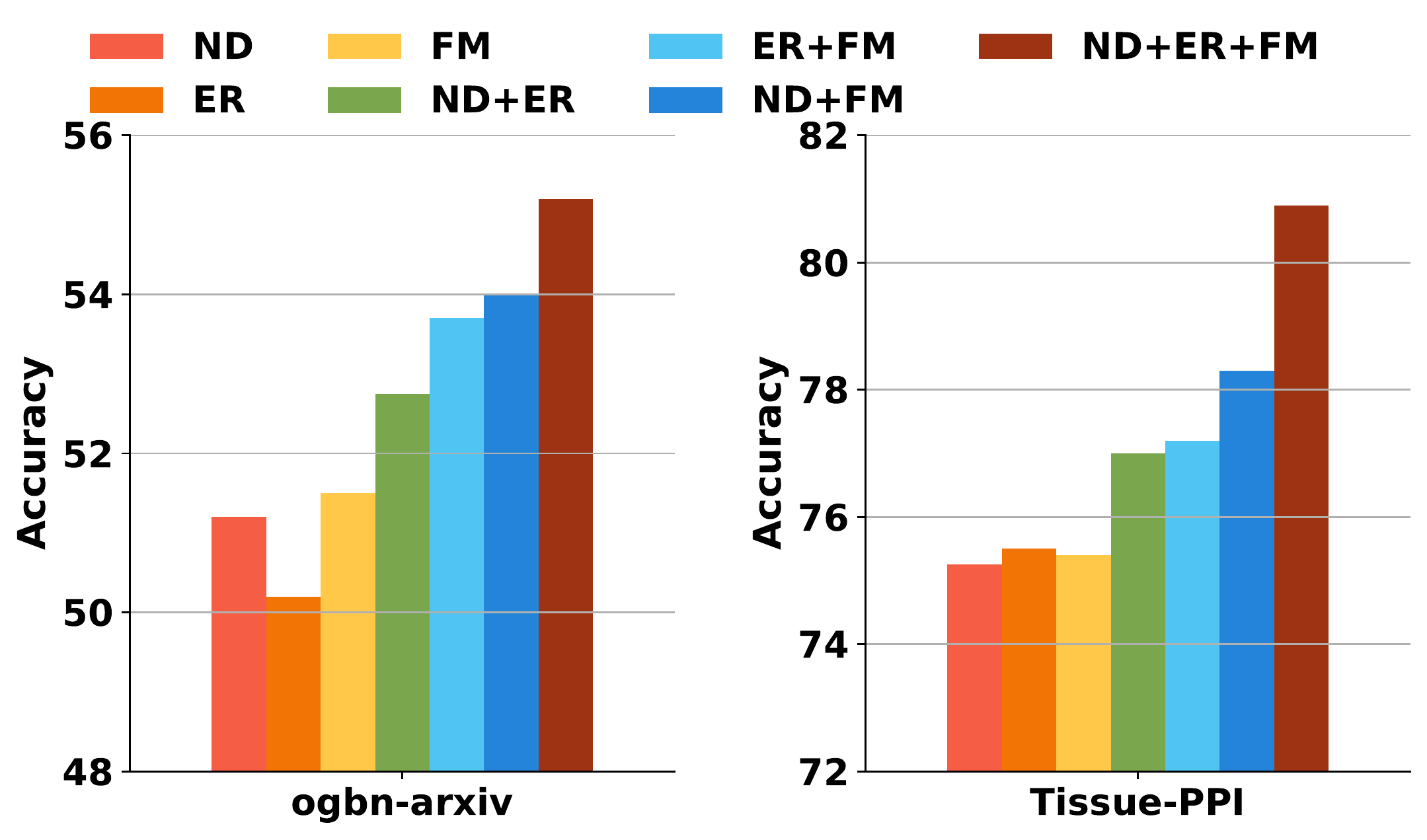}
\captionof{figure}{Impact of data aug. on node classification.}
\label{fig:node-da}
\end{minipage}
\hfill
\begin{minipage}{0.48\linewidth}
\begin{center}
\vspace{0.1in}
 \captionof{table}{Info. loss for node classification.}
\label{tab:info-loss-node-cls}
\scalebox{0.7}{
\setlength{\tabcolsep}{.6em}
        \begin{tabular}{c|c|c|c}
        \toprule
        Method &  Layer & ogbn-arxiv & Tissue-PPI\\
\midrule
                                     &GNN-1 &385.15 &788.30 \\
G-Meta                               &GNN-2 &383.39 &789.78 \\
                                     &FC  &374.42 &753.85\\
\midrule
                                     &GNN-1 &305.29 &663.38 \\
CGFL-I                              &GNN-2 &300.29 &650.59 \\
                                     &FC  &290.99 &612.98\\
\midrule
                                      &GNN-1 &\textbf{276.30} &\textbf{533.86}   \\
CGFL-T                               &GNN-2 &\textbf{267.33} &\textbf{525.42} \\
                                      &FC  &\textbf{264.23} &\textbf{503.86}  \\
\bottomrule
    \end{tabular}
}
\end{center}
\end{minipage}

\textbf{Loss Information Comparison}.
In Section~\ref{sec:inforamtion-loss}, we propose to compute graph information discarded in the GFL model. Here, we compare results between CGFL and a selected baseline method (G-Meta) in Table~\ref{tab:info-loss-node-cls}. From this table, the amount of discarded information in each layer of CGFL is smaller than baseline models. This may be because CGFL can learn more label-irrelevant information from unlabeled graph data, which somehow shows the superiority of CGFL in learning node embeddings for node classification.  

\vspace{-0.15in}
\subsection{Few-Shot Graph Classification}
\vspace{-0.1in}
\textbf{Overall Performance}.
The results of all models for 5/10-shot graph classification
are reported in Table~\ref{tab:graph-cls-1}. Similar to the findings obtained from Table~\ref{tab:node-cls-1}, according to this table: (1) CGFL outperforms all baseline models, showing its superiority for few-shot graph classification;
(2) The improvement of CGFL-I over baseline models ranges from 0.95\% to 38.36\% (5-shot) and from 0.8\% to 34.33\% (10-shot). Meanwhile, this value of CGFL-T ranges from 5.65\% to 46.21\% (5-shot) and from 2.18\% to 41.51\% (10-shot). This demonstrates the effect of contrastive pre-training and self-distillation in learning rich graph embedding from unlabeled data; (3) CGFL-T (or Teacher-T) outperforms CGFL-I (or Teacher-I), showing that unlabeled data in testing set improves model's generalization; {(4) CGFL is better than teacher model as contrastive distillation step further elevates the model.}

\begin{figure}
\centering
\begin{subfigure}{.5\textwidth}
  \centering
  \includegraphics[width=1\linewidth]{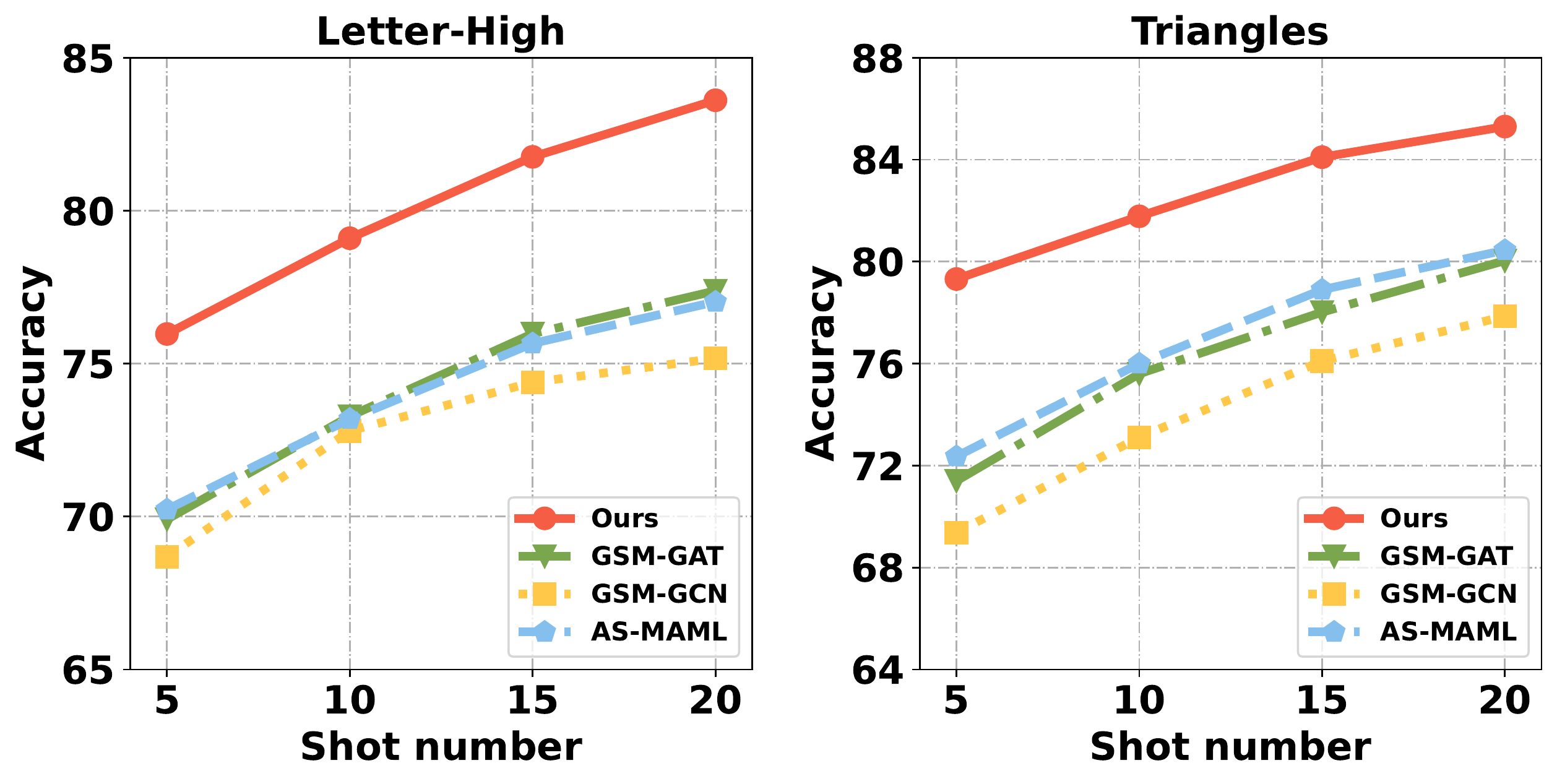}
  \caption{Impact of shot number.}
\label{fig:graph-shot}
\end{subfigure}%
\begin{subfigure}{.5\textwidth}
  \centering
  \includegraphics[width=1\linewidth]{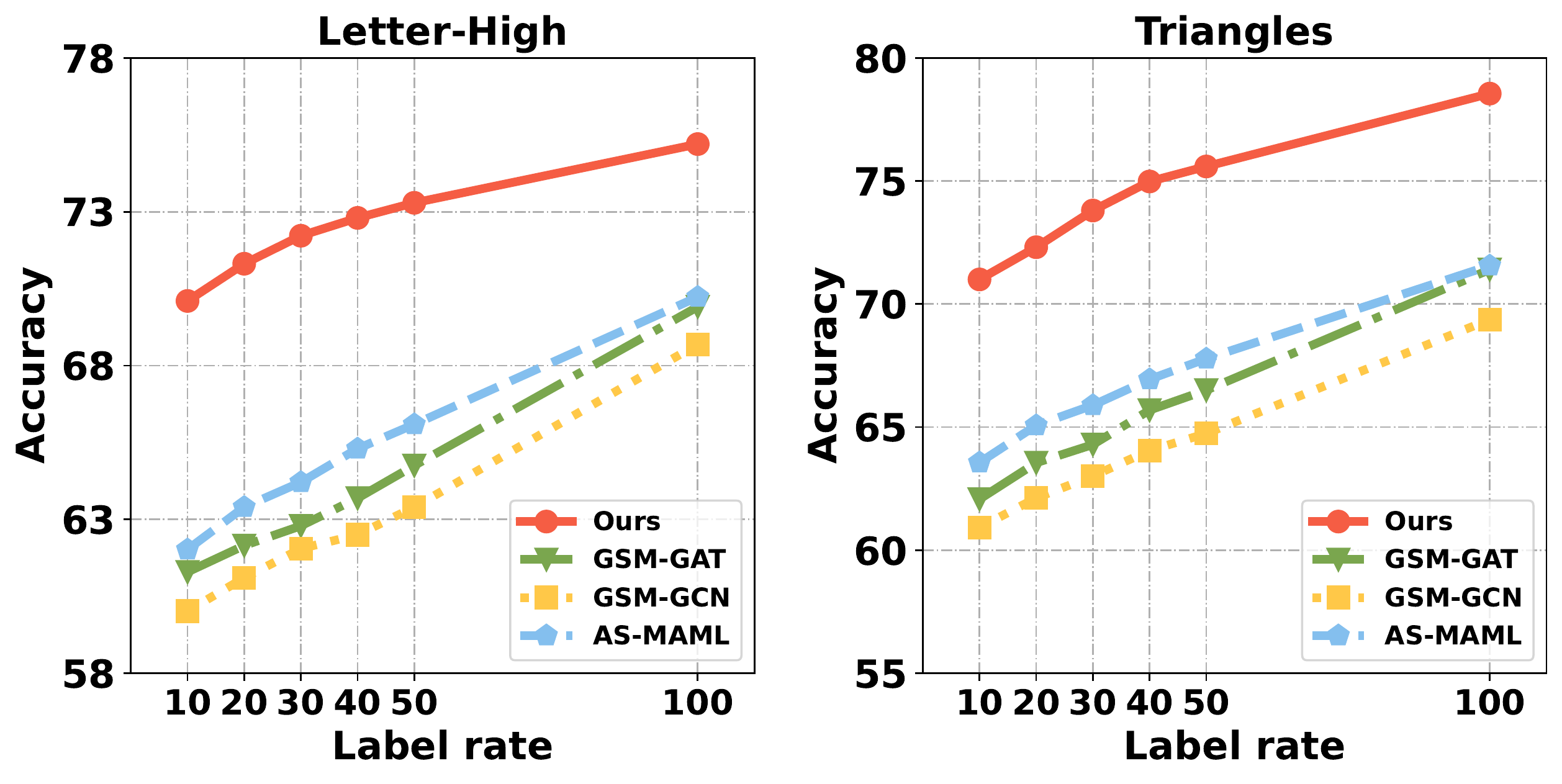}
  \caption{Impact of label rate.}
  \label{fig:graph-lr}
\end{subfigure}
\caption{Impact of shot number and label rate on graph classification.}
\vspace{-0.1in}
\label{fig:graph-shot-lr}
\end{figure}

\begin{table*}[t]
\vspace{-0.1in}
\caption{Few-shot graph classification results. The best results are highlighted in bold while the best baseline results are underlined. -I and -T denote inductive and transductive settings of CGFL, respectively. Teacher indicates the CGFL model without knowledge distillation.}
\vspace{-0.1in}
\label{tab:graph-cls-1}
\centering
\resizebox{1\textwidth}{!}{
\begin{tabular} {c|cc|cc|cc|cc}
\toprule
\multirow{2}{*}{Method} & \multicolumn{2}{c|}{Letter-High} &\multicolumn{2}{c|}{Triangles} &\multicolumn{2}{c|}{Reddit-12K} &\multicolumn{2}{c}{Enzymes}  \\ 
\cmidrule{2-9}
& 5-shot & 10-shot & 5-shot & 10-shot & 5-shot & 10-shot & 5-shot & 10-shot  \\
\midrule
WL &65.27$\pm$7.67&68.39$\pm$4.69  &51.25$\pm$4.02&53.26$\pm$2.95 &40.26$\pm$5.17 &42.57$\pm$3.69 &55.78$\pm$4.72 &58.47$\pm$3.84  \\ 
Graphlet &33.76$\pm$6.94&37.59$\pm$4.60  &40.17$\pm$3.18&43.76$\pm$3.09 &33.76$\pm$6.94 &37.59$\pm$4.60 &53.17$\pm$5.92 &55.30$\pm$3.78  \\ 
AWE &40.60$\pm$3.91&42.20$\pm$2.87  &39.36$\pm$3.85&42.58$\pm$3.11 &30.24$\pm$2.34 &33.44$\pm$2.04 &43.75$\pm$1.85 &45.58$\pm$2.11\\ 
Graph2Vec  &66.12$\pm$5.21&68.17$\pm$4.26  &48.38$\pm$3.85&50.16$\pm$4.15 &27.85$\pm$4.21 &29.97$\pm$3.17 &55.88$\pm$4.86 &58.22$\pm$4.30  \\ 
Diffpool  &58.69$\pm$6.39&61.59$\pm$5.21 &64.17$\pm$5.87&67.12$\pm$4.29 &35.24$\pm$5.69 &37.43$\pm$3.94 &45.64$\pm$4.56 &49.64$\pm$4.23 \\ 
CapsGNN  &56.60$\pm$7.86&60.67$\pm$5.24 &65.40$\pm$6.13&68.37$\pm$3.67 &36.58$\pm$4.28 &39.16$\pm$3.73 &52.67$\pm$5.51 &55.31$\pm$4.23  \\ 
GIN &65.83$\pm$7.17&69.16$\pm$5.14 &63.80$\pm$5.61&67.30$\pm$4.35 &40.36$\pm$4.69 &43.70$\pm$3.98 &55.73$\pm$5.80 &58.83$\pm$5.32 \\ 
GIN-KNN &63.52$\pm$7.27&65.66$\pm$8.69 &58.34$\pm$3.91&61.55$\pm$3.19 &41.31$\pm$2.84 &43.58$\pm$2.80 &57.24$\pm$7.06 &59.34$\pm$5.24 \\ 
GSM-GCN &68.69$\pm$6.50&72.80$\pm$4.12 &69.37$\pm$4.92&73.11$\pm$3.94 &40.77$\pm$4.32 &44.28$\pm$3.86 &54.34$\pm$5.64 &58.16$\pm$4.39 \\ 
GSM-GAT &69.91$\pm$5.90&\underline{73.28$\pm$3.46} &71.40$\pm$4.34 &\underline{75.60$\pm$3.67} &41.59$\pm$4.12 &\underline{45.67$\pm$3.68} &55.42$\pm$5.74 &60.64$\pm$3.84 \\ 
AS-MAML &\underline{70.23$\pm$1.53} &73.19$\pm$1.17 &\underline{71.56$\pm$1.17} &75.56$\pm$2.39 &\underline{41.90$\pm$1.65} &45.66$\pm$1.11 &\underline{56.03$\pm$1.85} &\underline{60.79$\pm$2.74} \\ 
\midrule
Teacher-I &{71.43$\pm$5.23} &{73.62$\pm$2.93}  &{71.93$\pm$3.51} &{76.21$\pm$2.87} &{42.32$\pm$4.48} &{46.31$\pm$3.84}  &{57.53$\pm$3.16} &{61.12$\pm$4.80} \\
CGFL-I &{72.12$\pm$4.88} &{74.08$\pm$3.30} &{72.34$\pm$3.42} &{76.91$\pm$2.98} &{42.85$\pm$4.62} &{46.89$\pm$4.96} &{57.94$\pm$2.85} &{61.66$\pm$4.61}\\
Teacher-T &{75.20$\pm$4.34} &{78.35$\pm$1.95}   &{78.55$\pm$3.75} &{81.03$\pm$3.37} &{44.80$\pm$4.85} &{48.95$\pm$4.03}  &{59.85$\pm$2.34} &{62.30$\pm$3.29}  \\
CGFL-T &\textbf{75.97$\pm$5.02} &\textbf{79.10$\pm$4.23}  &\textbf{79.32$\pm$4.05} &\textbf{81.78$\pm$3.30} &\textbf{45.55$\pm$3.67} &\textbf{49.32$\pm$4.21} &\textbf{60.34$\pm$4.04} &\textbf{62.97$\pm$2.92}  \\
\midrule
\end{tabular}}
\vspace{-0.2in}
\end{table*}
\textbf{Impact of Shot Number}.
In Figure~\ref{fig:graph-shot}, we report our model's result under different shot numbers (5, 10, 15, 20) compared with some selected baselines. Similar to Figure~\ref{fig:node-shot}, CGFL consistently outperforms baseline methods across different shot numbers, showing the robustness of CGFL. Note that we only show results on two datasets (e.g., Letter-High and Triangles), and the results on the other datasets are shown in Appendix~\ref{sec:app:fsg}.
The same goes for the following analyses.  
\\
\vspace{-0.1in}
\\
\textbf{Impact of Training Label Rate}.
In Figure~\ref{fig:graph-lr}, we report our model's performance under different training label rates compared with baseline models for 5-shot graph classification. The phenomenon is similar to the node task: 
our model achieves better accuracy across different label rates; the performance gaps between CGFL and baseline methods are larger when the label rate is lower. Again, this shows the significance of CGFL when labeled data is limited.  
\\
\textbf{Impact of Data Augmentation}.
We also study the impact of graph augmentation on few-shot graph classification task. 
According to Figure~\ref{fig:graph-da}, we find that the combination of three augmentation strategies brings the best performance, showing the importance of sufficient contrastive pairs on model training. 
\\
\begin{minipage}{0.52\textwidth}
\centering
\label{fig:graph-aug}
\vspace{0.1in}
    \includegraphics[width=0.9\linewidth]{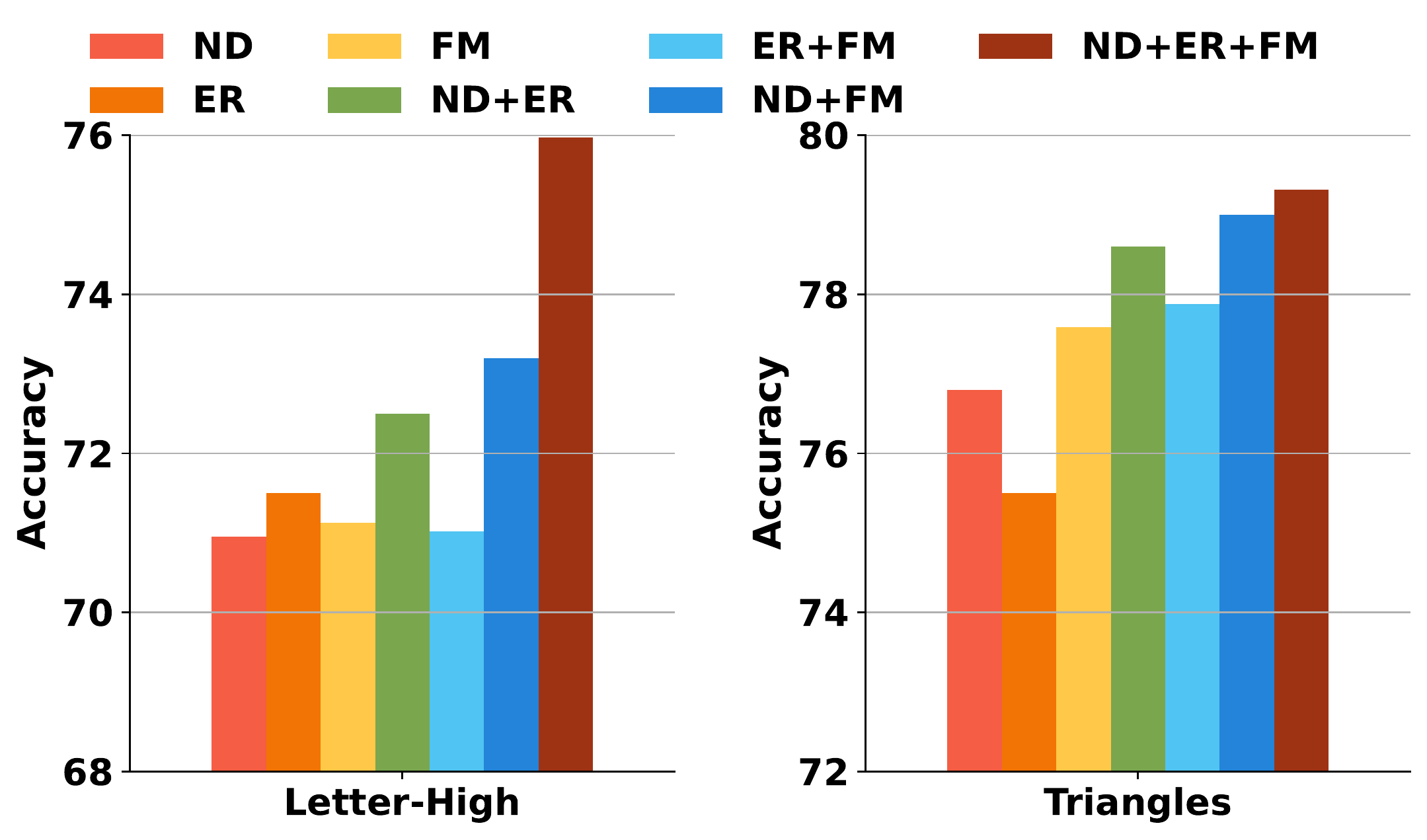}
\captionof{figure}{Impact of data aug. on graph classification.}
\vspace{0.1in}
\label{fig:graph-da}
\end{minipage}
\hfill
\begin{minipage}{0.48\linewidth}
\begin{center}
\vspace{0.1in}
\captionof{table}{Info. loss for graph classification.}
\label{tab:info-loss-graph-cls}
\scalebox{0.7}{
\setlength{\tabcolsep}{.6em}
        \begin{tabular}{c|c|c|c}
        \toprule
        Method &  Layer & Letter-High & Triangles\\
\midrule
      &GNN-1  &1286.67 &1520.94 \\
GSM-GAT &GNN-2 &1250.54 &1487.42 \\
      &FC   &1105.96 &1364.11 \\
\midrule
                 &GNN-1 &{932.23}  &{1276.77}  \\
CGFL-I  &GNN-2 & {920.58} & {1219.91} \\
                 &FC-3  & {892.03} &{1107.56}  \\
\midrule
                &GNN-1 &\textbf{792.65} &\textbf{923.35} \\
CGFL-T &GNN-2 &\textbf{780.59} &\textbf{894.09} \\
                &FC  &\textbf{765.73} &\textbf{885.35} \\
\bottomrule
    \end{tabular}
    \vspace{0.2in}
}
\end{center}
\end{minipage}

\textbf{Loss Information Comparison}.
Finally, we compute the discarded graph information of our model and a selected baseline method (GSM-GAT), as shown in Table~\ref{tab:info-loss-graph-cls}.
Obviously, the amount of information discarded in CGFL is less than baseline models. It somehow shows the superiority of CGFL in learning graph embeddings for graph classification.

\vspace{-0.1in}

\vspace{-0.1in}
\section{Conclusion}
\vspace{-0.1in}
\label{sec:conclusion}
In this paper, to tackle the limitations of existing GFL models, we propose a general and effective framework named CGFL - \textbf{C}ontrastive \textbf{G}raph \textbf{F}ew-shot \textbf{L}earning. CGFL leverages a self-distilled contrastive learning procedure to boost GFL, in which the GNN encoder is pre-trained with contrastive learning and further elevated with knowledge distillation in a self-supervised manner. Additionally, we introduce an information-based method to compute the amount of graph information discarded by the GFL model. Extensive experiments on multiple 
graph datasets demonstrate that CGFL outperforms state-of-the-art baseline methods for both node classification and graph classification tasks in the few-shot scenario. The discarded information value further shows the superiority of CGFL in learning node (or graph) embeddings. 


\clearpage
\bibliographystyle{plain}

\clearpage
\appendix
\maketitle
\section{Dataset Details}
\label{app: dataset}
For few-shot node classification, we use five different datasets: ogbn-arxiv~\cite{hu2020open}, Tissue-PPI~\cite{hamilton2017inductive}, Fold-PPI~\cite{zitnik2017predicting}, Cora~\cite{sen2008collective}, and Citeseer~\cite{sen2008collective} to perform extensive empirical evaluations of different models. These datasets  vary from citation network to biochemical graph. The statistics of datasets are reported in Table~\ref{tab:node-data-stat}. 

\begin{table}[htbp]
\vspace{-0.1in}
\caption{Statistics of datasets used in the node classification task.}
\label{tab:node-data-stat}
\vspace{1mm}
\small
\renewcommand\arraystretch{1}
\centering
\resizebox{0.67\textwidth}{!}{\begin{tabular*}{0.7\textwidth}{c| @{\extracolsep{\fill}} |c|c|c|c|c}
\toprule
{Dataset} &  \# Graph & \# Node & \#  Edge &  \# Feat. &  \# Label  \\
\midrule
ogbn-arxiv       &1 &169,343 &1,166,243 &128 &40  \\ 
Tissue-PPI       &24 &51,194 &1,350,412 &50 &10 \\ 
Fold-PPI       &144 &274,606 &3,666,563 &512 &29 \\ 
Cora       &1 &2,708 &10,556 &1,433 &7 \\ 
Citeseer       &1 &3,327 &9,228 &3,703 &6 \\ 
\bottomrule
\end{tabular*}
}
\end{table}

For few-shot graph classification, we use four different datasets~\cite{chauhan2020few}: Reddit-12K, ENZYMES, Letter-High, and TRIANGLES, to perform extensive empirical evaluations of different models. These datasets vary from small average graph size (e.g., Letter-High) to large graph size (e.g., Reddit-12K). 
The statistics of datasets are reported in Table~\ref{tab:graph-data-stat}. 
\begin{table}[htbp]
\vspace{-0.1in}
\caption{Statistics of datasets used in the graph classification task.}
\label{tab:graph-data-stat}
\vspace{1mm}
\small
\renewcommand\arraystretch{1}
\centering
\resizebox{0.6\textwidth}{!}{\begin{tabular*}{0.6\textwidth}{c| @{\extracolsep{\fill}} |c|c|c|c|c}
\toprule
\multirow{2}{*}{Dataset} & \multicolumn{2}{c|}{Class \#} &\multicolumn{3}{c}{Graph \#}  \\
\cmidrule{2-3} \cmidrule{4-6} &Train &Test & Training &Validation &Test  \\
\midrule
Letter-High       &11 &4 &1,330 &320 &600 \\ 
Triangles       &7 &3 &1,126 &271 &603 \\ 
Reddit-12K       &7 &4 &566 &141 &404 \\ 
Enzymes       &4 &2 &320 &80 &200 \\ 
\bottomrule
\end{tabular*}
}
\end{table}

\section{Baseline Method Details}
\label{app: implementation}
\subsection{Node Classification}
\label{sec:app:fsn}

\textbf{Graph embedding models:}

\textit{node2vec~\cite{grover2016node2vec}}: We use node2vec to generate node embeddings, then employ a FC layer as a predictor to classify nodes. We use the code at this link\footnote{https://shorturl.at/sEINW}.

\textit{DeepWalk~\cite{perozzi2014deepwalk}}: Similar to node2vec, we use DeepWalk to generate node embeddings, then employ an FC layer as a predictor to classify nodes. We use the code at this link\footnote{https://github.com/phanein/deepwalk}.

\textbf{GNN-based models:}

\textit{{Meta-GNN~\cite{zhou2019meta}:}} It combines MAML and simple graph convolution (SGC) to learn node embeddings. We use the code at this link\footnote{https://github.com/ChengtaiCao/Meta-GNN}.

\textit{{FS-GIN~\cite{xu2018powerful}:}} This method uses GIN to learn node embeddings and only uses few-shot nodes to propagate loss and enable training. We use the code for GIN backbone at this link\footnote{https://github.com/weihua916/powerful-gnns\label{footnote:gin}}.

\textit{{FS-SGC~\cite{wu2019simplifying}:}} This model is similar to FS-GIN while changing GIN to SGC as GNN backbone. We use the code of SGC at this link\footnote{https://github.com/Tiiiger/SGC}.

\textit{{No-Finetune~\cite{huang2020graph}:}} This method trains a GCN on the support set and uses the trained backbone to classify samples in the meta-testing set. We use the code of GCN at this link\footnote{https://shorturl.at/lwFPR\label{footnote:ft}}.

\textit{{Finetune~\cite{triantafillou2019meta}:}} This method trains GCN on the meta-training set, and the model is fine-tuned on the meta-testing set. We use the code of GCN at this link$^{\ref{footnote:ft}}$.

\textit{{KNN~\cite{triantafillou2019meta}:}} This method trains a GNN on meta-training set. Then, it uses the label of the K-closest examples in the support set for each query example. We use the related code at this link\footnote{https://shorturl.at/etBO8}.

\textit{{ProtoNet~\cite{triantafillou2019meta}:}} This method applies prototypical network on node embeddings processed by a neural network, which is trained under the standard meta-learning setting. We use the related code at this link\footnote{https://shorturl.at/erQU9}.

\textit{{MAML~\cite{finn2017model}:}} It is similar to ProtoNet  but changes meta-learner from ProtoNet to MAML. We use the code at this link\footnote{https://github.com/mims-harvard/G-Meta\label{footnote:gmeta}}.

\textit{{G-Meta~\cite{huang2020graph}:}} This is state-of-the-art baseline for few-shot node classification. It uses GCN as GNN backbone to learn node embeddings based on local subgraphs. It further combines prototypical loss and  MAML for model training. We use the code at this link$^{\ref{footnote:gmeta}}$.

\textit{{TENT~\citep{wang2022task}:}} This is also a state-of-the-art baseline for few-shot node classification. It proposes task-adaptive node classification framework to make node-level, class-level, and task-level adaptations. We utilize the code at this link.\footnote{https://github.com/SongW-SW/TENT}

\subsection{Graph Classification}
\label{sec:app:fsg}
\textbf{Graph embedding models:}

\textit{{WL~\cite{shervashidze2011weisfeiler}:}} It uses KNN search on the output embeddings of WL. We use the code of WL at this link\footnote{https://github.com/BorgwardtLab/P-WL}.

\textit{{Graphlet~\cite{shervashidze2009efficient}:}} It uses Graphlet Kernel to decompose a graph and generates graph embeddings. We use the code of Graphlet at this link\footnote{https://github.com/paulmorio/geo2dr\label{footnote:geo2dr}}.

\textit{{AWE~\cite{ivanov2018anonymous}:}} It uses KNN search on the output embeddings of AWE. We use the code of AWE at this link$^{\ref{footnote:geo2dr}}$.

\textit{{Graph2Vec~\cite{narayanan2017graph2vec}:}} This method applies KNN search on the output embeddings of Graph2Vec. We use the code of Graph2Vec at this link$^{\ref{footnote:geo2dr}}$.

\textbf{GNN-based models:}

\textit{{Diffpool~\cite{lee2019self}:}} It uses Diffpool with supervised loss to generate graph embeddings. We use the code of Diffpool at this link\footnote{https://github.com/RexYing/diffpool}.

\textit{{CapsGNN~\cite{xinyi2018capsule}:}} This method applies CapsGNN to generate graph embeddings with supervised training. We use the code of CapsGNN backbone at this link\footnote{https://github.com/benedekrozemberczki/CapsGNN}.

\textit{{GIN~\cite{xu2018powerful}:}} This model applies GIN to generate graph embeddings with supervised training. We use the code of GIN backbone at this link$^{\ref{footnote:gin}}$.

\textit{{GIN-KNN~\cite{xu2018powerful}:}} Similarly, this model implements GIN to generate graph embeddings while it switches the MLP classifier to the KNN algorithm. We use the code of GIN backbone at this link$^{\ref{footnote:gin}}$.

\textit{{GSM-GCN~\cite{chauhan2020few}:}} This is a state-of-the-art model (with GCN as backbone) for few-shot graph classification. We follow the default settings in the original paper and use the code at this link\footnote{https://github.com/chauhanjatin10/GraphsFewShot\label{footnote:gsm}}.

\textit{{GSM-GAT~\cite{chauhan2020few}:}} This is a state-of-the-art model (with GAT as backbone) for few-shot graph classification. We follow the default settings in the original paper and use the code at this link$^{\ref{footnote:gsm}}$.

\textit{{AS-MAML~\cite{ma2020adaptive}:}} It is a state-of-the-art model for few-shot graph classification.  We follow the default settings in the original paper and use the code at this link\footnote{https://github.com/NingMa-AI/AS-MAML}.

\section{Additional Experiment Results}
\label{app: results}
This section shows additional results on the other three datasets (i.e., Fold-PPI, Cora, and Citeseer) for few-shot node classification and the other two datasets (i.e., Reddit-12K and Enzymes) for few-shot graph classification. 

\subsection{Few-Shot Node Classification Results}
\begin{figure}[!htb]
\centering
\includegraphics[width=1.0\linewidth]{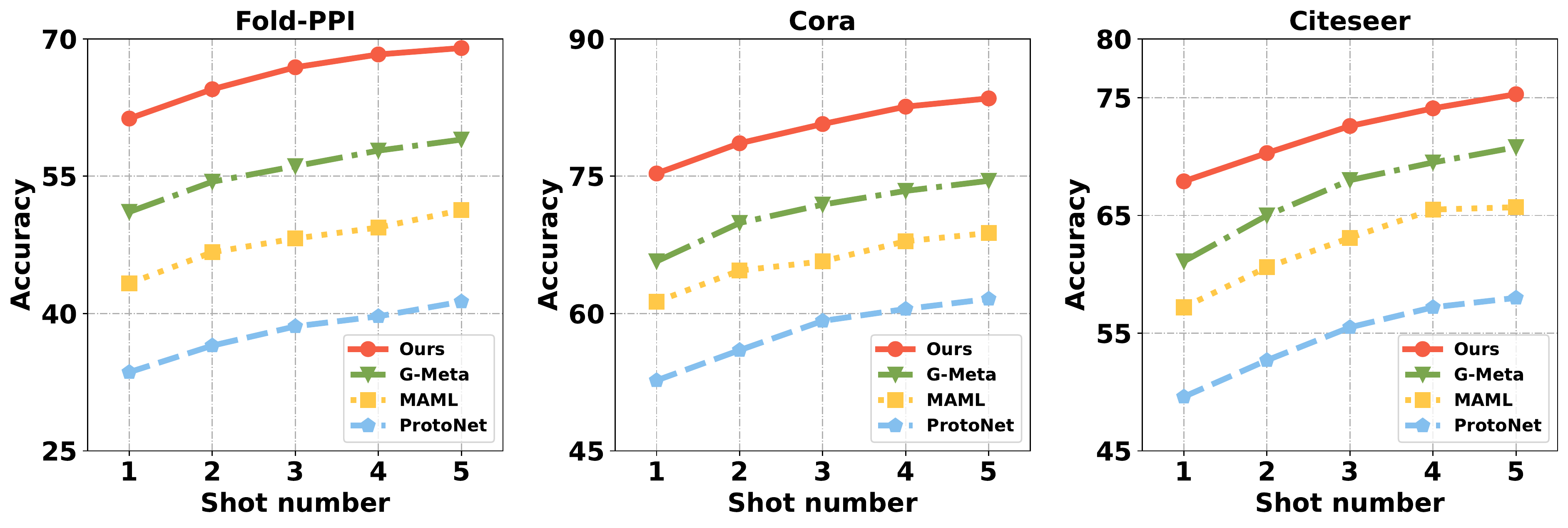}
\vspace{-0.2in}
\caption{Impact of shot number on node classification.}
\vspace{-0.2in}
\label{fig:app-node-shot}
\end{figure}

\textbf{Impact of Shot Number}. Figure~\ref{fig:app-node-shot} shows our model's performance under different shot numbers (1 to 5) compared with some selected baselines. It is easy to see that CGFL consistently outperforms baseline methods across different shot numbers.

\begin{figure}[!htb]
\centering
\includegraphics[width=1.0\linewidth]{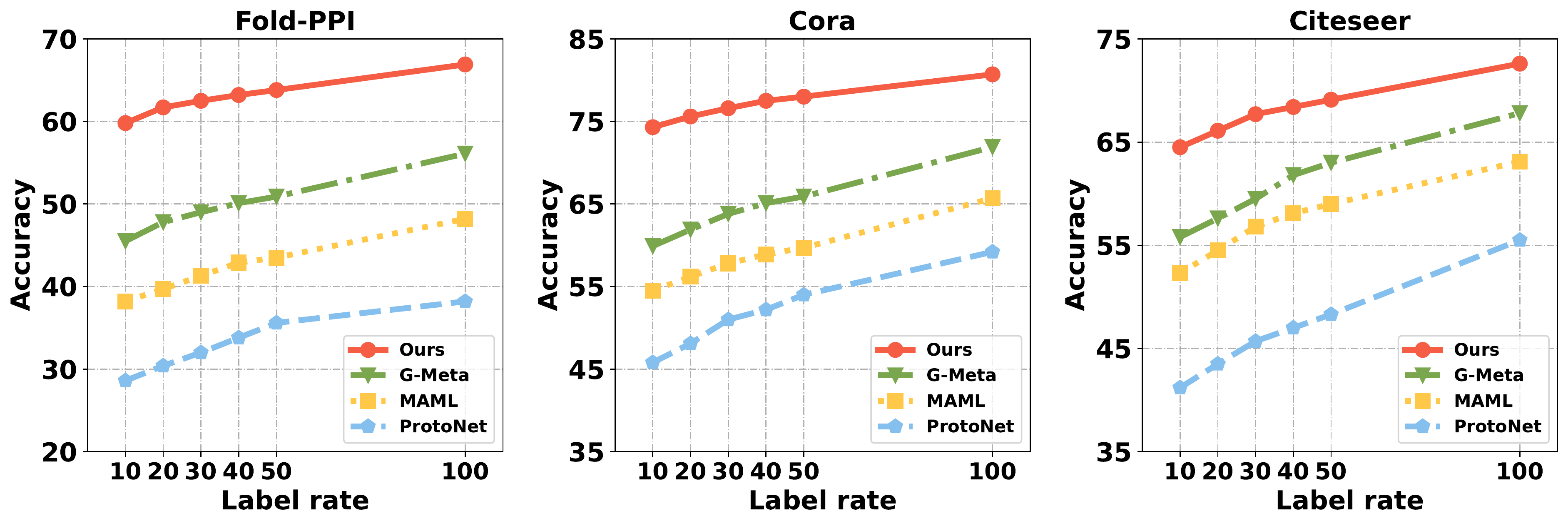}
\vspace{-0.25in}
\caption{Impact of training label rate on node classification.}
\vspace{-0.15in}
\label{fig:app-node-lr}
\end{figure}

\textbf{Impact of Training Label Rate}. Figure~\ref{fig:app-node-lr} reports our model's performance under different training label rates compared with baseline models for 3-shot node classification. Obviously, our model achieves better accuracy across different label rates.

\begin{figure}[!htb]
\centering
\includegraphics[width=0.8\linewidth]{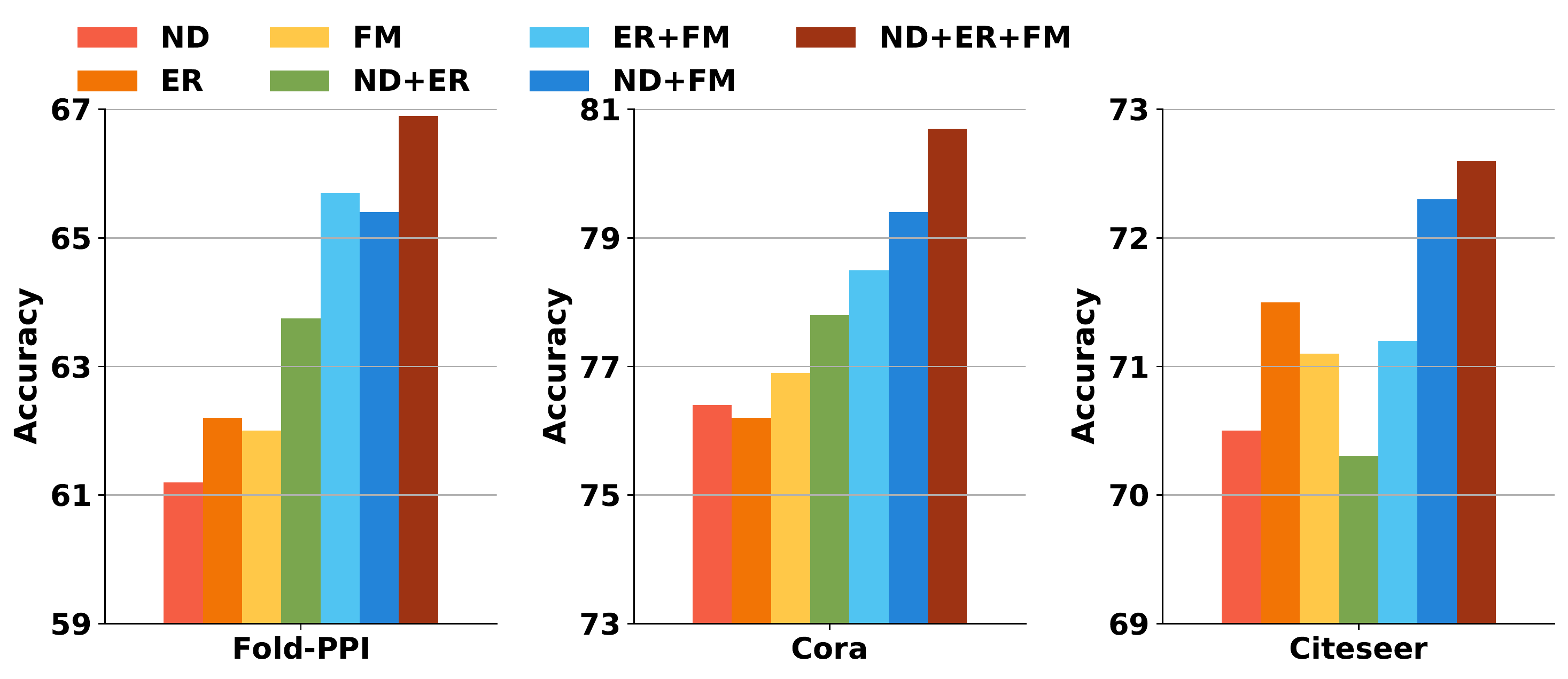}
\caption{Impact of data augmentation on node classification.}
\label{fig:app-node-da}
\end{figure}

\textbf{Impact of Data Augmentation}. The impact of graph augmentation on node classification is shown in Figure~\ref{fig:app-node-da}. According to this figure, we can find that the combination of three augmentation strategies brings the best performance.

\textbf{Discarded Information Comparison}. The discarded graph information of different models are shown in Table~\ref{tab:info-loss-node-cls-app}.
Obviously, the amount of information discarded in CGFL is less than baseline models. 
\begin{table}[htbp]
\caption{Discarded information for node classification.}
\small
\centering
\resizebox{0.57\textwidth}{!}{\begin{tabular*}{0.55\textwidth}{c|c|c|c|c}
\toprule
\multirow{2}{*}{Method} & \multirow{2}{*}{Layer} &\multicolumn{3}{c}{Discarded Information}  \\ 
\cmidrule{3-5} & & Fold-PPI &Cora &Citeseer \\%
\midrule
                                     &GNN-1 &618.52  &361.70 &134.20 \\ 
Meta-GNN                             &GNN-2  &612.11  &357.47 &126.75 \\ 
                                     &FC   &591.13 &327.49 &116.55 \\ \midrule
                                     &GNN-1  &585.31 &356.57 &129.04 \\ 
G-Meta                               &GNN-2  &592.38  &355.63 &119.72 \\ 
                                     &FC   &553.25   &318.87 &100.10 \\ 
\midrule
                                     &GNN-1  &509.10  &340.72 &111.90 \\ 
CGFL-I                              &GNN-2  &511.10  &326.35 &105.01 \\ 
                                     &FC     &461.52  &316.21 &92.95 \\ 
\midrule
                                      &GNN-1   &\textbf{426.23} &\textbf{332.39} &\textbf{104.04} \\ 
CGFL-T                               &GNN-2   &\textbf{418.90}  &\textbf{314.65} &\textbf{88.33} \\ 
                                      &FC    &\textbf{384.23}  &\textbf{306.98}  &\textbf{78.25}\\ 
\bottomrule
\end{tabular*}
}
\label{tab:info-loss-node-cls-app}
\end{table}

\subsection{Few-Shot Graph Classification Results}
\begin{figure}[!htb]
\centering
\includegraphics[width=0.8\linewidth]{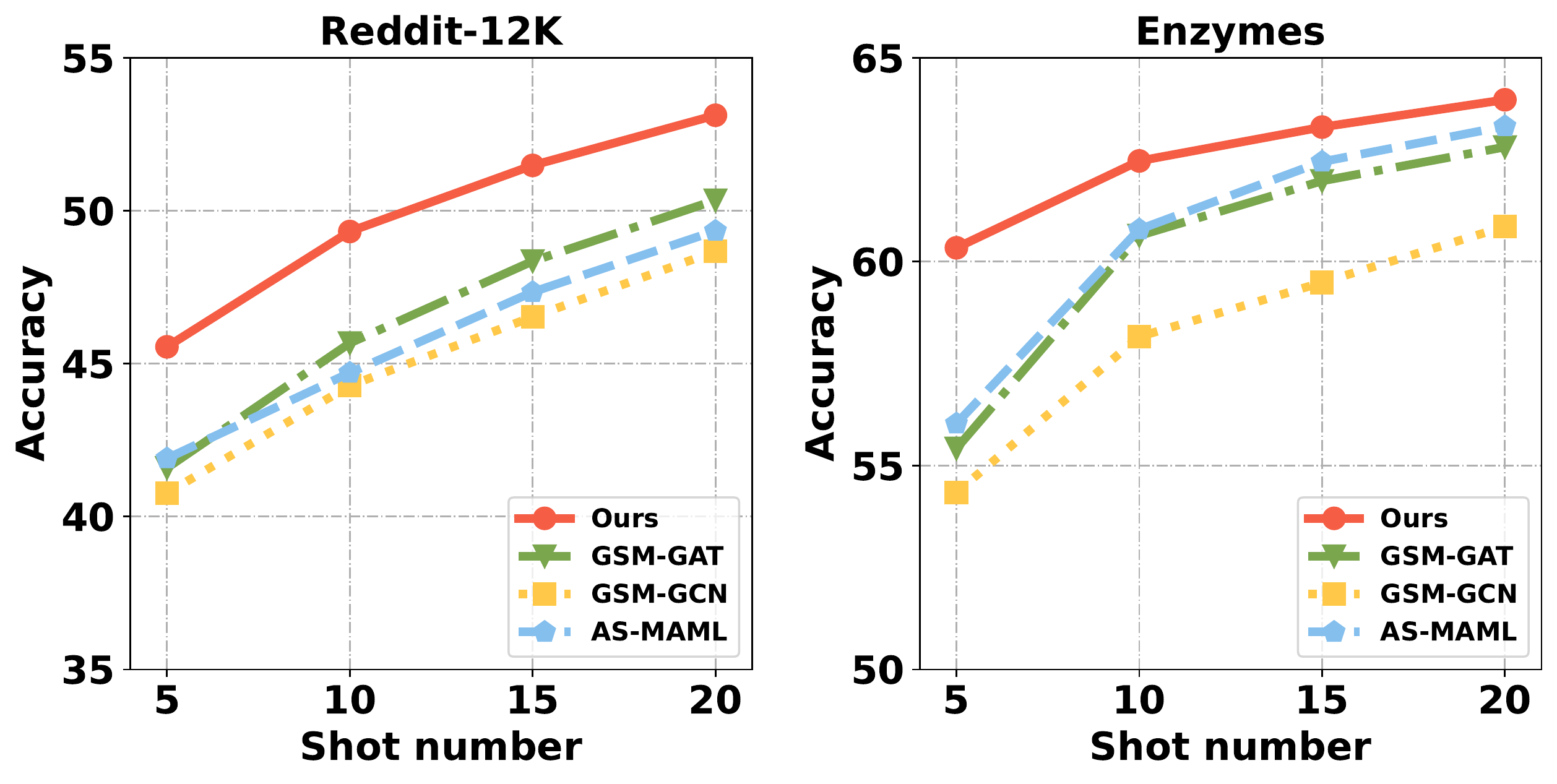}
\caption{Impact of shot number on graph classification.}
\label{fig:app-graph-shot}
\end{figure}

\textbf{Impact of Shot Number}. Figure~\ref{fig:app-graph-shot} shows CGFL's performance under different shot numbers (5, 10, 15, 20) compared with some selected baselines. From this figure, our model consistently outperforms baseline methods across different shot numbers. 

\begin{figure}[!htb]
\centering
\includegraphics[width=0.8\linewidth]{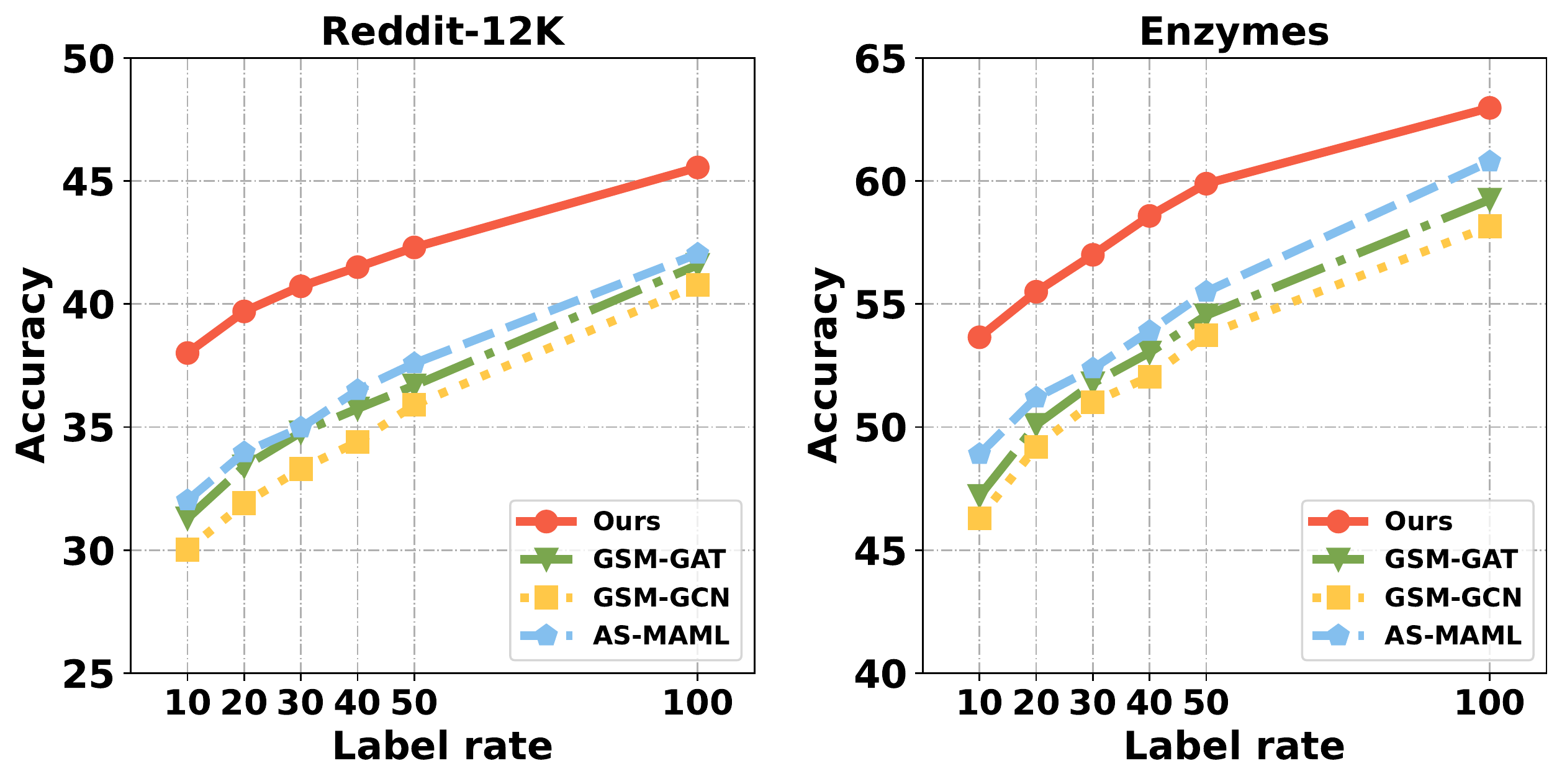}
\vspace{-0.1in}
\caption{Impact of training label rate on graph classification.}
\label{fig:app-graph-lr}
\end{figure}

\textbf{Impact of Training Label Rate}. Figure~\ref{fig:app-graph-lr} shows our CGFL's result under different training label rates compared with baseline models for 5-shot graph classification. Obviously, CGFL achieves better performance across different label rates.  

\begin{figure}[t]
\centering
\includegraphics[width=0.6\linewidth]{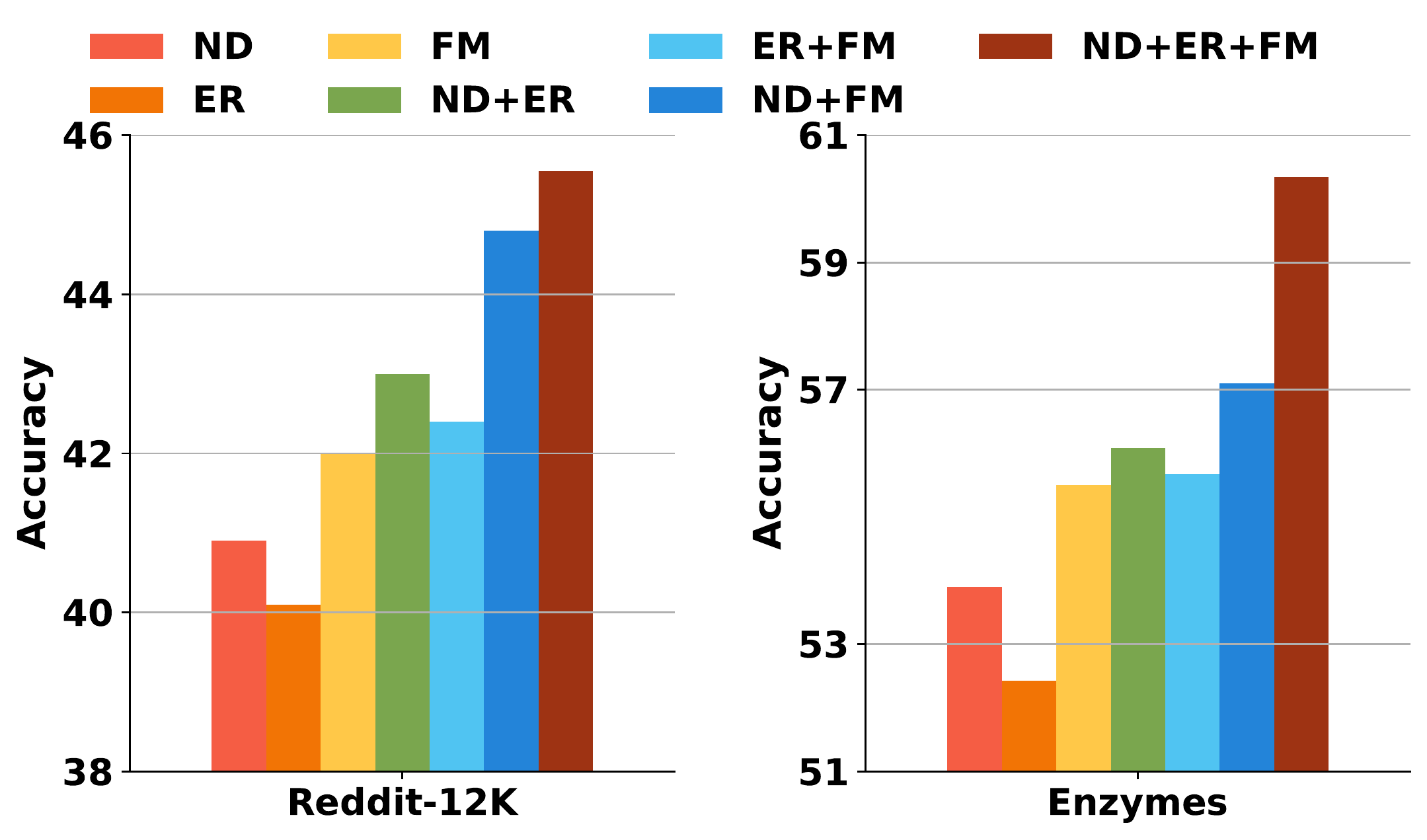}
\vspace{-0.1in}
\caption{Impact of data augmentation on graph classification.}
\vspace{-0.1in}
\label{fig:app-graph-da}
\end{figure}

\textbf{Impact of Data Augmentation}. The impact of graph augmentation on graph classification is shown in Figure~\ref{fig:app-graph-da}. From this figure, we can see that the combination of three augmentation strategies leads to the best result.

\textbf{Discarded Information Comparison}. The discarded graph information of different methods are reported in Table~\ref{tab:info-loss-graph-cls-app}.
According to this table, the amount of information discarded in CGFL is less than baseline models. 

\begin{table}[!htb]
\caption{{Discarded information for graph classification.}}
\vspace{1mm}
\small
\centering
\resizebox{0.5\textwidth}{!}{\begin{tabular*}{0.5\textwidth}{c|c|c|c}
\toprule
\multirow{2}{*}{Method} & \multirow{2}{*}{Layer} &\multicolumn{2}{c}{Discarded Information}  \\ 
\cmidrule{3-4} & &Reddit-12K &Enzymes \\%
\midrule
       &GNN-1   &5455.46 &2455.54 \\
AS-MAML &GNN-2 &5024.15  &2243.25 \\
       &FC     &4937.99 &2125.91\\
       \midrule
       &GNN-1  &5401.30  &2273.80 \\ 
GSM-GAT &GNN-2 &5183.87  &2128.04 \\ 
       &FC     &4936.76  &1958.18\\ 
\midrule
                &GNN-1  &5323.04  &1864.98 \\ 
CGFL-I         &GNN-2  &5098.54  &1788.97 \\ 
                &FC     &4802.03  &1759.87 \\  
\midrule
                 &GNN-1 &\textbf{4823.91}  &\textbf{1787.98} \\ 
CGFL-T          &GNN-2 &\textbf{4546.23}  &\textbf{1698.35} \\ 
                   &FC  &\textbf{4329.39}  &\textbf{1585.33} \\ 
\bottomrule
\end{tabular*}
}
\label{tab:info-loss-graph-cls-app}
\end{table}

\subsection{{Apply to MAML++ and ALFA in Few-shot Learing Phase}}
{We implement MAML++ and ALFA (extended from MAML) in our CGFL to replace MAML under transductive setting on ogbn-arxiv dataset. The results are reported in  Table~\ref{tab:node-alfa}. Our aim is to design a contrastive pretraining framework which is plug-and-play for graph few-shot learning. So our CGFL is not limited to MAML. It is easy to be extended with more new methods (e.g., MAML++). Replacing MAML to MAML++ in CGFL brings better accuracy, which implies our framework is general and extendable.}
\begin{table}[t]
\caption{{Comparison with finetuning by MAML++ and ALFA.}}
\label{tab:node-alfa}
\vspace{1mm}
\small
\renewcommand\arraystretch{1}
\centering
\resizebox{0.4\textwidth}{!}{\begin{tabular*}{0.45\textwidth}{c| @{\extracolsep{\fill}} |c|c}
\toprule
{Model} &  3-shot & 5-shot \\
\midrule
MAML++       &55.9$\pm$2.1 &59.4$\pm$3.2 \\
ALFA       &54.8$\pm$3.6 &58.8$\pm$3.7 \\
Ours (with MAML)       &55.2$\pm$2.5 &58.7$\pm$2.7  \\
\bottomrule
\end{tabular*}
}
\end{table}

\subsection{{Apply to Few-shot Image Classification}}
{We have made an attempt to transfer our method to the few-shot image classification task. Specifically, we use ResNet-12 to get image features. The results of our method and a representative GNN-based method~\cite{yang2020dpgn, tang2021mutual} on inductive mini-ImageNet 5-way classification are reported in Table~\ref{tab:gnn-based}. It demonstrates that our method can outperform the selected baseline method and has good potential in few-shot image classification task.}
\begin{table}[!htb]
\caption{{Comparison with GNN-based method for few-shot image dataset mini-ImageNet.}}
\label{tab:gnn-based}
\small
\renewcommand\arraystretch{1}
\centering
\resizebox{0.4\textwidth}{!}{\begin{tabular*}{0.45\textwidth}{c| @{\extracolsep{\fill}} |c|c}
\toprule
{Model} &  1-shot    & 5-shot \\
\midrule
DPGN~\cite{yang2020dpgn} &57.04±0.52 &72.83±0.74 \\
MCGN~\cite{tang2021mutual} &57.89±0.87 &73.58±0.87\\
Ours    & 60.02±0.54 & 76.45±0.88\\
\bottomrule
\end{tabular*}
}
\end{table}

\section{{Clarification about Related Work}}
{Both motivation and method design of our self-distillation strategy are different from Fixmatch~\cite{sohn2020fixmatch}: (a) Our CGFL is motivated by graph few-shot learning problem where label domain distribution shift exists between training data and test data. Thus, we use contrastive self-supervised training to learn more label-irrelevant representation for alleviating label domain distribution shift issue. It is different from FixMatch’s aim to learn from partially labeled images; (b) During contrastive pre-training and self-distillation procedures, our model learns representation without labeled graph data while Fixmatch requires partially labeled images during contrastive training. }


\end{document}